\documentclass[runningheads]{llncs}
\let\llncssubparagraph\subparagraph
\let\subparagraph\paragraph

\usepackage{eccv}

\usepackage{eccvabbrv}

\usepackage{graphicx}
\usepackage{booktabs}
\usepackage{multirow}
\usepackage[export]{adjustbox}
\usepackage{wrapfig}
\usepackage{titlesec}
\let\subparagraph\llncssubparagraph
\usepackage{bm}
\usepackage{subcaption,siunitx,booktabs}

\titlespacing{\paragraph}{0pt}{2pt}{4pt}
\titlespacing{\section}{0pt}{10pt}{3pt}
\titlespacing{\subsection}{0pt}{5pt}{2pt}

\usepackage[accsupp]{axessibility}  %

\newcommand{\ourmethod}{ProCreate}
\newcommand{\ourdataset}{FSCG-8}

\usepackage[colorlinks=true]{hyperref}

\usepackage{orcidlink}
\usepackage{tabu}
\usepackage{xurl}

\begin{document}

\title{
ProCreate, Don't Reproduce!\newline Propulsive Energy Diffusion for\newline Creative Generation
} 

\titlerunning{Propulsive Energy Diffusion for Creative Generation}

\author{Jack Lu \and
Ryan Teehan \and
Mengye Ren}

\authorrunning{J.~Lu \etal}

\institute{New York University\\ \email{\{yl11330,rst306,mengye\}@nyu.edu}
\\
Project Webpage: \url{https://procreate-diffusion.github.io}
}

\maketitle

\begin{abstract}
In this paper, we propose \ourmethod{}, a simple and easy-to-implement method to improve sample diversity and creativity of diffusion-based image generative models and to prevent training data reproduction. \ourmethod{} operates on a set of reference images and actively propels the generated image embedding away from the reference embeddings during the generation process. We propose FSCG-8 (Few-Shot Creative Generation 8), a few-shot creative generation dataset on eight different categories---encompassing different concepts, styles, and settings---in which \ourmethod{} achieves the highest sample diversity and fidelity. Furthermore, we show that \ourmethod{} is effective at preventing replicating training data in a large-scale evaluation using training text prompts. Code and \ourdataset{} are available at \url{https://github.com/Agentic-Learning-AI-Lab/procreate-diffusion-public}.
\end{abstract}

\begin{figure}[h!]
  \vspace{-0.4in}
  \centering
  \includegraphics[trim={0cm 0cm 0cm 0cm},clip,width=\textwidth,center]{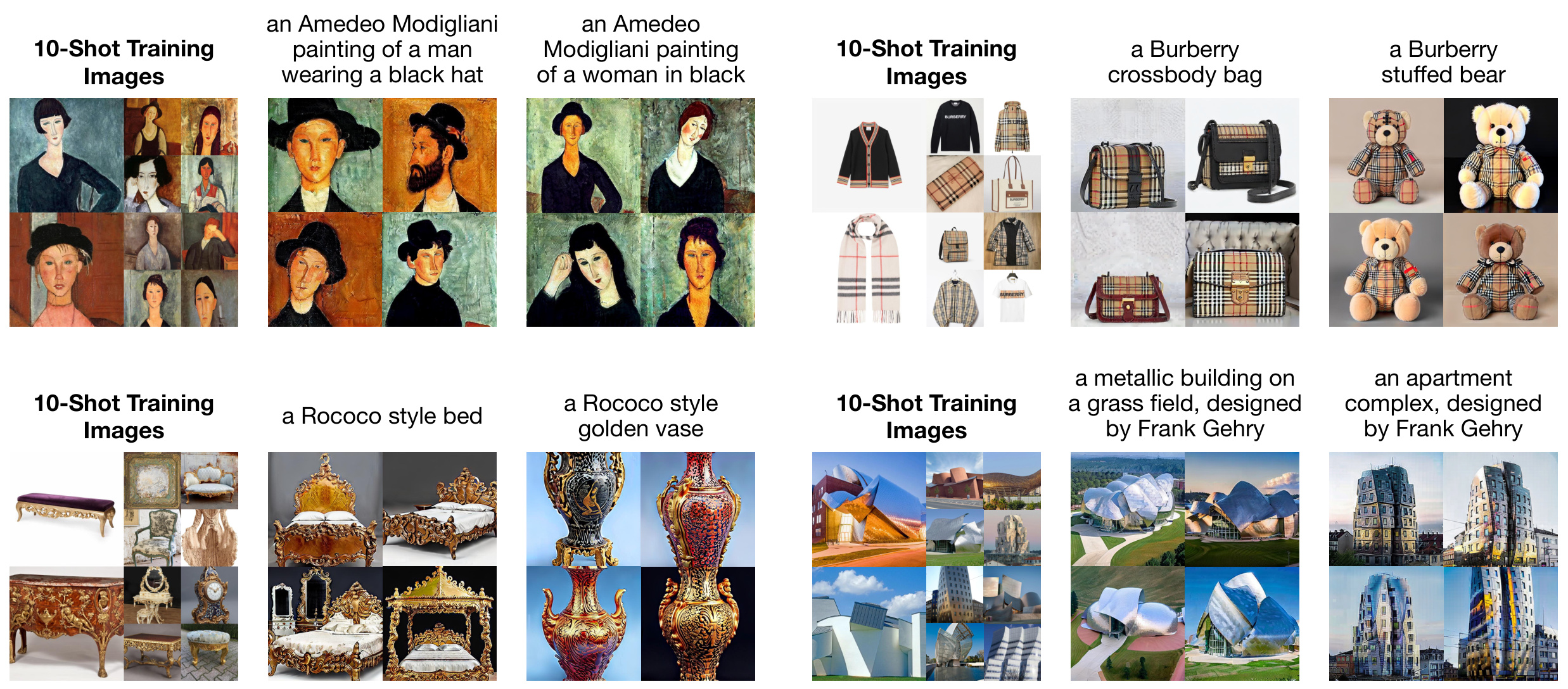}
  \caption{After fine-tuning a diffusion model on each category of our few-shot dataset \ourdataset{}, \emph{\ourmethod{}} can significantly improve the diversity and creativity of generations while retaining high image quality and prompt fidelity. 
  }
  \label{fig:teaser}
\end{figure}

\section{Introduction}\label{sec:introduction}

\label{sec:intro}
Imagine a fashion designer attempting to brainstorm a new idea for a clothing line. Looking back on past runway shows that they found particularly stunning and influential, they attempt to draw inspiration from others' work without copying it directly. They want to evoke similar concepts, expressed perhaps in the silhouettes and proportions showcased on the runway, the particular way the fabric drapes on each model's frame, or the drama communicated by a specific cut or color palette. In other words, their goal is to take a reference set of images, which express a unique creative vision from a designer, and draw inspiration from them without direct reproduction. If they attempt to use a generative image model for this creative iteration, however, they would find that the model had either never acquired that ineffable concept before, and thus was unable to produce good examples, or had memorized examples from the reference set during adaptation (e.g. through fine-tuning or tuning new token embeddings \cite{gal2022textual_inversion}), and merely reproduces elements of the design exactly. Our paper aims to address this problem directly, propelling generated images away from those in a reference set while maintaining high-level conceptual inspiration. In that way, we \textit{jointly} address the specific problem of exact replication \textit{and} the general problem of low diversity among generative model samples.

Recent advances in generative image modeling, in particular the development of highly capable, and easier to train, denoising diffusion probabilistic models (DDPMs) \cite{ho2020denoising}, have enabled high-quality, complex image generation in both conditional (e.g. class- or text-conditional generation) and unconditional settings \cite{dhariwal2021diffusion, saharia2022imagen, nichol2021glide,rombach2022latent_diffusion,ramesh2022dalle2}. With their widespread adoption, diffusion models are increasingly used in customized settings, where a smaller number of images are used to define a particular domain \cite{zhu2023domainstudio, oneshotdomainadaptation}, style \cite{specialistdiffusion, disentangledstyle, inversionstyle}, subject/concept \cite{ruiz2023dreambooth,gal2022textual_inversion,customdiffusion}.

Unfortunately, despite their improvements over GANs, diffusion models are similarly prone to training data memorization and a lack of sample diversity \cite{somepalli2023diffusion_copying, somepalli2023diffusion_forgery, particleguidance}, which is particularly damaging in light of their use in these customized, low-data settings---generated images are often similar to one another \cite{sadat2023cads}. This behavior is particularly damaging when diffusion models are used to produce art or graphic design materials, where creativity is essential, since the models may replicate potentially copyrighted examples directly~\cite{somepalli2023diffusion_forgery,somepalli2023diffusion_copying}, opening users and model developers up to potential liability. 

To improve sampled image diversity, we propose an energy-based \cite{Hinton2000TrainingPO, song2021train, du2023reduce, variationaldiffusion, scoresde} method which applies a propulsive force to latent representations during inference, pushing the generated image away from the images in the reference set. We consider two experimental setups: 1) few-shot creative generation and 2) training data replication prevention. For few-shot generation, we construct a new few-shot image generation dataset called \ourdataset{} using collected images across eight different categories. While fine-tuning a diffusion model to a reference set quickly overfits in this low-data setting, \ourmethod{}, by contrast, achieves greater sample diversity while retaining broad conceptual similarity, both being essential components of creative generation. In the training data replication experiments, we show that \ourmethod{} is effective at preventing pre-trained diffusion models from replicating training data.

In summary, our contributions are as follows:
\begin{itemize}
    \item To test sample diversity, memorization, and creative generation, we collected a few-shot generation dataset{}, \ourdataset{}, spanning across domains such as paintings, architecture, furniture, fashion, cartoon characters, etc. 
    \item We propose \ourmethod{}, a simple and easy-to-implement component that allows diffusion models to generate diverse and creative images using concepts from a reference set, without direct reproduction.
    \item In few-shot generation, \ourmethod{} is found to have better sample diversity than prior arts, while maintaining a high similarity to the reference set.
    \item \ourmethod{} addresses the training data replication issue with a significantly lower chance of replicating training images than pre-trained diffusion models.
\end{itemize}

\section{Related Work}\label{sec:related_work}
\looseness=-10000
In this section, we review related areas of work in guided diffusion models, few-shot image generation, sample diversity, and data replication issues.

\paragraph{Guided Diffusion Models.} Conditioning provides a powerful and flexible way to guide diffusion model generation using text \cite{saharia2022photorealistic, rombach2022latent_diffusion}, images \cite{lugmayr2022repaint, ho2022cascaded, palette}, videos \cite{videodiffusion,imagenvideo}, layout \cite{layout}, or even scene graphs \cite{scenegraph1,scenegraph2} and point clouds \cite{lion,pointclouddiffusion}. Recent works have explored classifier-free guidance and classifier guidance as two methods for class- or text-conditional image generation \cite{dhariwal2021diffusion,classifierguidance}. Classifier-free guidance requires training a new DDPM that accepts an additional conditioning input, but achieves strong performance for conditional image generation tasks, such as conditioning on class \cite{dhariwal2021diffusion}, prompts or regions of the original image \cite{nichol2021glide}, among others. On the other hand, classifier guidance avoids re-training by guiding a pre-trained model using a classifier to guide the sampling process. Recent work applied classifier guidance across a variety of conditioning goals without training new classifiers \cite{bansal2023universal}. DOODL \cite{doodl} improves the performance of off-the-shell classifier guidance by performing backpropagation through the entire diffusion inference process to optimize the initial noise.

\paragraph{Few-Shot Image Generation.} Diffusion models can be adapted to a few-shot setting \cite{giannone2022few}, including for customization and personalization \cite{gal2022textual_inversion, ruiz2023dreambooth}. Giannone \etal \cite{giannone2022few} developed a method for few-shot adaptation of diffusion models, but they focused on CIFAR-100~\cite{Krizhevsky2009LearningML}, which only contains $32\times 32$ images. In contrast, we collect our own few-shot generation dataset with $512\times 512$ images, where we choose each class to be practical for designers and for the images of each class to contain more conceptual similarities (e.g., Burberry design) than simply belonging to the same semantic category (e.g., horse). Standard fine-tuning, Dreambooth \cite{ruiz2023dreambooth}, and Textual Inversion \cite{gal2022textual_inversion} all allow for customizing diffusion models. The first two methods require fine-tuning an entire diffusion model to produce images with a consistent subject, which makes it prone to overfitting. Textual Inversion learns a new token for the concept it wants to replicate, which can have a more limited capacity for representing novel concepts from new training images. Since \ourmethod{} is applicable to any diffusion model, it can easily be applied on top of either of these approaches. At the same time, neither method handles issues with memorization and sample diversity, whereas \ourmethod{} can adapt to the concept(s) in the few-shot reference set without direct reproduction or low sample diversity.

\paragraph{Sample Diversity.}
\looseness=-10000
CADS \cite{sadat2023cads} is recently proposed as a method to address the sample diversity problem in diffusion models. CADS uses annealed conditioning during inference, allowing them to balance the quality and diversity of samples from the model. In contrast, \ourmethod{} guides the generated sample to be different from the set of training images, achieving even better sample diversity than CADS and simultaneously avoiding reproducing potentially copyrighted images in the training set. Sehwag \etal \cite{sehwag2022generating} introduces a framework that involves sampling from low-density regions of the data manifold to avoid reproducing training images and generating novel samples. However, their method operates in pixel space rather than in latent space like ours, making it difficult to apply to latent diffusion models. Finally, we explore the low-data few-shot setting where one wants to not only avoid reproduction but also draw inspiration from the limited training images, while CADS and Sehwag \etal \cite{sehwag2022generating} do not.

\paragraph{Data Replication.}
\looseness=-10000
We also draw on prior work studying data replication in generative models, particularly diffusion models. Carlini \etal \cite{carlini2023extracting} constructed a pipeline that is used to extract thousands of training examples from state-of-the-art diffusion models, Somepalli \etal \cite{somepalli2023diffusion_forgery} studied the rate at which diffusion models' inference outputs replicate their training data at various data sizes, and Somepalli \etal \cite{somepalli2023diffusion_copying} proposed de-duplicating images in the training set and randomizing/augmenting captions to reduce the rate of output replications. Similar issues were studied in GANs both theoretically \cite{Nagarajan2019Theoretical_memorization_gans} and empirically \cite{bai2022reducing_memorization_gans, bai2021training}. In contrast to Somepalli \etal \cite{somepalli2023diffusion_copying}, \ourmethod{} addresses the issue of diffusion output replication more directly by guiding each generation away from a set of images (e.g., all training images) we want to avoid generating.

\section{Background}\label{sec:background}

\looseness=-10000
In this section, we will cover the basics of diffusion denoising probabilistic models (DDPM) \cite{ho2020denoising}, denoising diffusion implicit models (DDIM) \cite{ddim}, and classifier guidance \cite{dhariwal2021diffusion}. Given samples from a data distribution $q(\bm{x}_0)$ of images, we can use diffusion models to learn a model distribution
$p_\theta(\bm{x}_0)$ that approximates $q(\bm{x}_0)$ and can be sampled from. To sample a new image, diffusion denoising probabilistic models (DDPMs) \cite{ho2020denoising} start with a sample of Gaussian noise $\bm{x}_T\sim \mathcal{N}(\bm{0}, \textbf{I})$, then repeatedly denoise $\bm{x}_t$ into $\bm{x}_{t-1}$ where $t$ decrements from $T$ to $1$. In this paper, we use a more performant sampling method than the default DDPM approach, termed denoising diffusion implicit models (DDIM) \cite{ddim}, which was shown to beat GAN models with only 25 sampling steps \cite{dhariwal2021diffusion}. At each denoising step with a noisy image $\bm{x}_t$, the DDIM sampling process first makes a one-step prediction of $\bm{\hat{x}}_0$ with Equation \ref{eq:onestep}, then it denoises $\bm{x}_t$ into $\bm{\hat{x}}_{t-1}$ by Equation \ref{eq:DDIM}. $\bm{\epsilon}_\theta$ represents the learned diffusion model with weights $\theta$ and $\alpha_t$ are scalar parameters that define the diffusion noise schedule.
\begin{align}\label{eq:onestep}
    \bm{\hat{x}}_0 &= \frac{\bm{x}_t - \sqrt{1 - \alpha_t}\bm{\epsilon}_\theta(\bm{x}_t, t)}{\sqrt{\alpha_t}},\\
    DDIM(\bm{x}_t, t) &= \sqrt{\alpha_{t-1}}\bm{\hat{x}}_0 + \sqrt{1 - \alpha_{t-1}}\bm{\epsilon}_\theta(\bm{x}_t, t).
\label{eq:DDIM}
\end{align}

To generate samples from diffusion models that are conditioned on ImageNet class labels, Dhariwal \etal \cite{dhariwal2021diffusion} introduced classifier guidance. They used the gradient of a noise-aware ImageNet classifier $f_\phi(y|\bm{x}_t)$ to perturb the noise prediction of each denoising step. Following the formulation in Bansal \etal \cite{bansal2023universal}, classifier-guidance of class label $c$ is applied by modifying the diffusion process with an additive gradient term:
\begin{equation}\label{eq:guide}
    \bm{\epsilon}' = \bm{\epsilon}_\theta(\bm{x}_t, t) + \sqrt{1 - \alpha_{t-1}}\nabla_{\bm{x}_t} l_{ce}(c, f_{cl}(\bm{x}_t)),
\end{equation}
where $l_{ce}$ is the cross-entropy loss and $f_{cl}$ is a noise-aware classifier, then replacing $\bm{\epsilon}_\theta(\bm{x}_t, t)$ with $\bm{\epsilon}'$ in Equation \ref{eq:DDIM}. Furthermore, Ho and Salimans~\cite{classifierguidance} formulated diffusion models as score-matching models and showed that Equation \ref{eq:guide} works by lowering the energy of data for which the classifier $f_{cl}$ predicts to be of class $c$ with high likelihood. Note that here, lower energy corresponds to a higher probability of being sampled. In the next section, we will use the notations and basic concepts and notations from here to develop \ourmethod{}.

\section{ProCreate: \\Propulsive Energy Diffusion for Creative Generation}\label{sec:method}

\begin{figure}[!t]
  \centering
  \includegraphics[trim={0cm 0.4cm 0cm 0.2cm},clip,width=\textwidth,center]{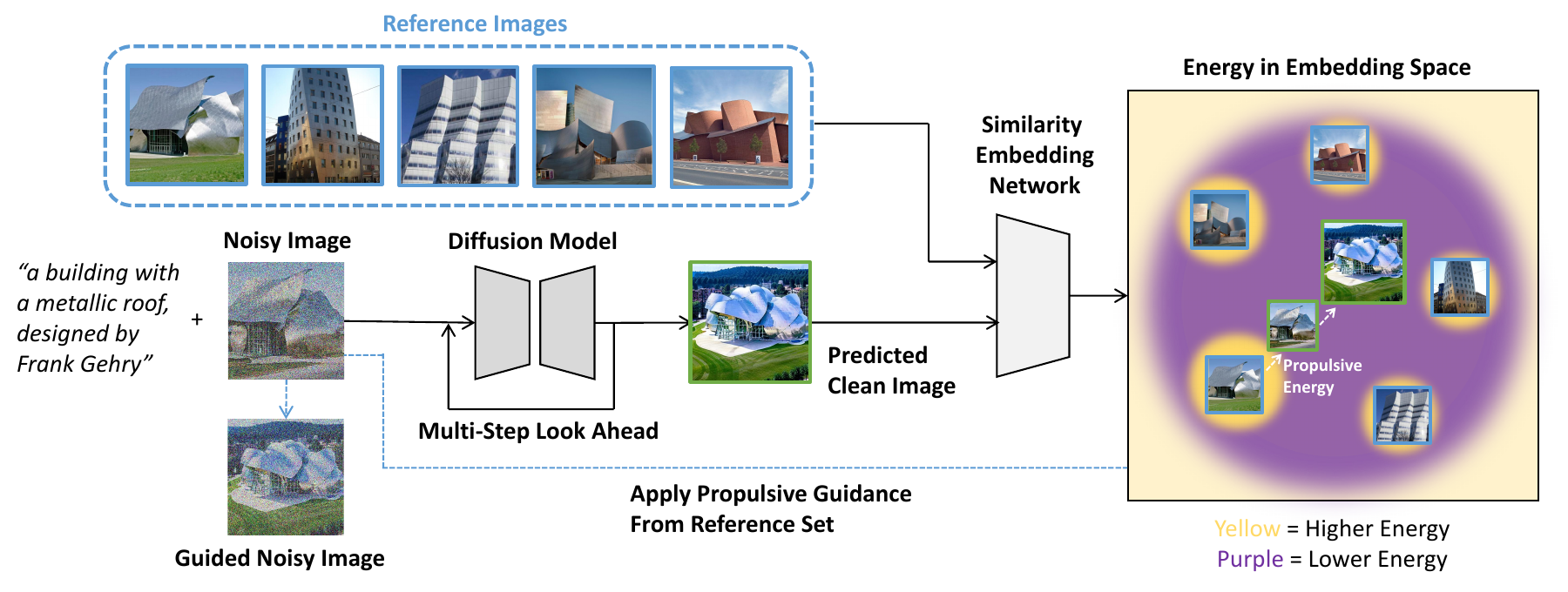}
  \caption{Overview of our approach. At each denoising step, \ourmethod{} applies gradient guidance that maximizes the distances between the generated clean image and the reference images in the embedding space of a similarity embedding network. In the embedding space, the noisy image is propelled away from its closest reference image.}
  \label{fig:main}
  \vspace{-0.2in}
\end{figure}

In this section, we introduce the mathematical formulation of \ourmethod{} and the techniques we use to improve \ourmethod{}'s performance. 
The core idea behind \ourmethod{} is to guide the generation away from images in the reference set using the gradient of an energy function computed with that set. 
As illustrated in Figure \ref{fig:main}, at each denoising step $t$ with the current noisy image $\bm{x}_t$, \ourmethod{} uses the diffusion model to predict a clean image $\bm{x}_0$, compute distances from $\bm{x}_0$ to each reference image, then finally updates $\bm{x}_t$ with a guidance gradient to maximize the distance between $\bm{x}_0$ with the closest reference image to it. With the guided noisy version of $\bm{x}_t$, \ourmethod{} can resume the normal diffusion model sampling process to denoise it to $\bm{x}_{t-1}$.

\paragraph{Energy Formulation.} Mathematically, we start with a reference set of $m$ images: $X^{r} = \{\bm{x}^{r}_{1}, \dots, \bm{x}^{r}_{m}\}$. Typically, they are either in the fine-tuning set for few-shot generation or the pre-training set. To mitigate the tendency of models to replicate these images, we guide the denoising process away using a gradient step from the log energy function that detects similarity among these images:
\begin{equation}\label{eq:procreateguide}
    \bm{\epsilon}' = \bm{\epsilon}_\theta(\bm{x}_t, t) + \sqrt{1 - \alpha_{t-1}}\nabla_{\bm{x}_t} g(\bm{x}_t, \{\bm{x}^r_1, \dots, \bm{x}^r_m \}),
\end{equation}
where the $g$ is defined as the log energy, which is the similarity between the predicted clean image $\hat{\bm{x}}_0$ and the closest reference image in the embedding space by using an embedding function $f$:
\begin{align}
    g(\bm{x}_t, \{\bm{x}_1, \dots, \bm{x}^r_m\}) &= \gamma \max_{i=1\dots m} s( f(\hat{\bm{x}}_0(\bm{x}_t)), f(\bm{x}^r_i) ).
\end{align}
\looseness=-10000
In practice, we use the cosine similarity for $s$, and we use DreamSim~\cite{fu2023dreamsim} as our embedding function $f$ since the DreamSim network is already pre-trained on detecting similar replicated images. $\gamma$ controls the strength of our energy function guidance. Note that the predicted clean image $\hat{\bm{x}}_0$ is a function of $\bm{x}_t$ that is predicted by Multi-Step Look Ahead, which we will explain below.

\paragraph{Multi-Step Look Ahead Prediction.} 
Referring to Section \ref{sec:background}, $\hat {\bm{x}}_0$ can be predicted in one-step by Equation~\ref{eq:onestep}. However, the one-step prediction of $\hat x_0$ at denoising steps when $t$ is large can be very inaccurate. Following DOODL~\cite{doodl}, to improve the quality of our estimate for $\hat x_0$ at each denoising step and in turn the quality of guidance, we perform DDIM diffusion $n_{step}$ times when predicting $\hat x_0$. Specifically, at denoising step $t$ with a noisy sample $x_t$, the Multi-Step Look Ahead prediction performs the following operations:

{
\tt
\hspace{15 mm} \noindent $\bm{x}\gets \bm{x}_t$

\hspace{15 mm} for $t'$ in $n_{step}$ timesteps evenly spaced from $t$ to $1$:

\hspace{25 mm} $\bm{x} \gets DDIM(\bm{x}, t')$

\hspace{15 mm} $\bm{\hat{x}}_0 \gets \bm{x}$
}

\paragraph{Dynamically Growing Reference Set.} To further improve the diversity of our generated samples, we add newly generated samples to the reference set after each batch $X^b=\{\bm{x}^{b,i}_0\}$ is generated:
$
X^{r} \gets X^{r} \cup X^b.
$
Therefore, the reference set continuously grows as we generate more samples, and new samples are guided away from the old ones to prevent diffusion models from generating similar images to previous ones.

\section{Few-Shot Creative Generation}\label{sec:fscg}

Given a diffusion model checkpoint that is fine-tuned on limited data, our goal is to improve the diversity of its generated samples while maintaining sample quality and conceptual similarity to the reference set. In this section, we describe our experiments on few-shot creative generation, comparing the samples generated from the default DDIM, CADS, and \ourmethod{} sampling methods both qualitatively and quantitatively.

\subsection{Our Dataset: \ourdataset{}}\label{subsec:our_dataset}

\begin{figure}[t]
  \centering
  \includegraphics[trim={0cm 0cm 0cm 0cm},clip,width=\textwidth,center]{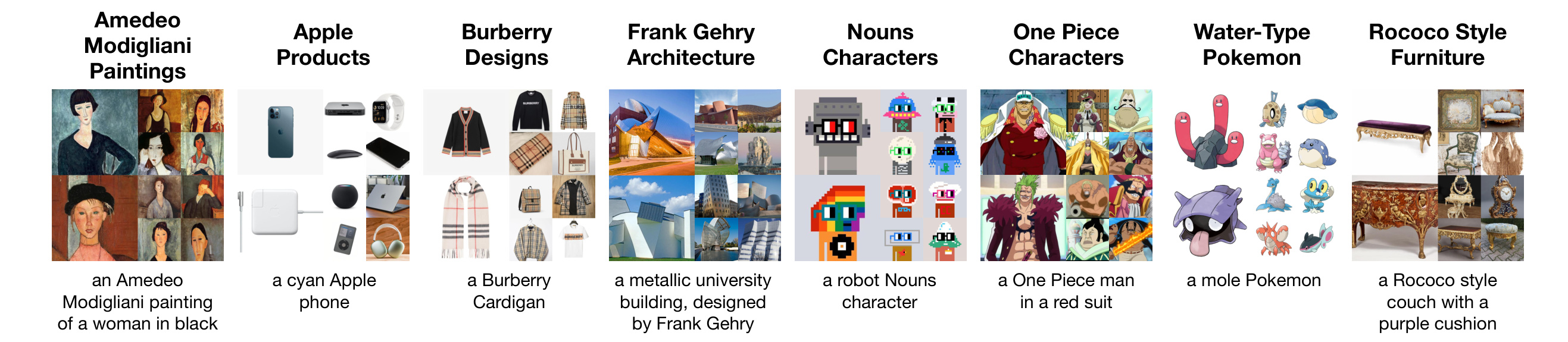}
  \caption{\looseness=-10000 Samples of the FSCG-8 dataset. We provide $10$ samples from each of the $8$ categories. For each category, an example caption for its top left image is provided.}
  \label{fig:datasets}
  \vspace{-0.25in}
\end{figure}

Our goal is to improve the performance of diffusion models on any few-shot training set with general shared properties, which can be subject, style, texture, and more. Diffusion models have been applied to a variety of few-shot learning tasks, Textual Inversion \cite{gal2022textual_inversion} and Dreambooth \cite{ruiz2023dreambooth} learn specific subjects, DomainStudio \cite{zhu2023domainstudio} performs domain adaptation, and Specialist Diffusion \cite{specialistdiffusion} learns styles, but none of these works curate datasets that share general properties. Therefore, we curate a new dataset FSCG-8 (Few-Shot Creative Generation 8), containing 8 categories of images with 50 prompt-image pairs in each. Samples of the dataset are shown in Figure~\ref{fig:datasets}.
Each category contains images that share some properties, including the style of Amedeo Modigliani's paintings, the abstract design and texture of Burberry's apparel, and the twisting geometric properties of Frank Gehry's architecture. We manually collect all images from the public domain on the Internet. For each category, we also design the prompts to be simple so that when given a validation prompt to a model, it has a large creative space of images to explore that all follow the prompt.

\subsection{Experiment Setup}\label{sec:fscg_experiment_setup}

\paragraph{Fine-tuning.} 
\looseness=-10000
We split each dataset in \ourdataset{} into $10$ training images and $40$ validation images then fine-tune a pre-trained Stable Diffusion v1-5 checkpoint with batch size 8 and learning rate $5\times 10^{-6}$ with no learning rate warm-up. We perform fine-tuning with two different training methods: standard fine-tuning for $2000$ iterations and DreamBooth fine-tuning for $6000$ iterations \cite{ruiz2023dreambooth}. DreamBooth fine-tuning is performed with prior preservation and we substitute the {[V]} token for class nouns ``Amedeo'', ``Apple'', ``Burberry'', ``Frank Gehry'', ``Nouns'', ``One Piece'', ``Pokemon'', and ``Rococo'' for each dataset in the order of Figure \ref{fig:datasets} respectively. For prior preservation, we set its loss weight to $0.5$, 
generate $100$ training images from prompts with the class noun removed, and split each batch such that half of the captions contain class nouns and the other half do not.

\paragraph{Inference.}
\looseness=-10000
For each trained checkpoint, we compare the quality and diversity of sampled images with DDIM, CADS, and \ourmethod{}. For each evaluation run, we generate $40$ samples from the $40$ prompts in the validation split. We set the number of inference steps to $50$ for all sampling methods. We tune hyperparameters $t_2$, $s$ for CADS and tune $\gamma$, gradient norm clipping (on the gradient of the energy function) for \ourmethod{}. For \ourmethod{}, we set the reference set to the training set and $n_{step}=5$ for Multi-Step Look Ahead.

\begin{figure}[!h]
  \centering
  \includegraphics[trim={0cm 0cm 0cm 0cm},width=\textwidth,center]
  {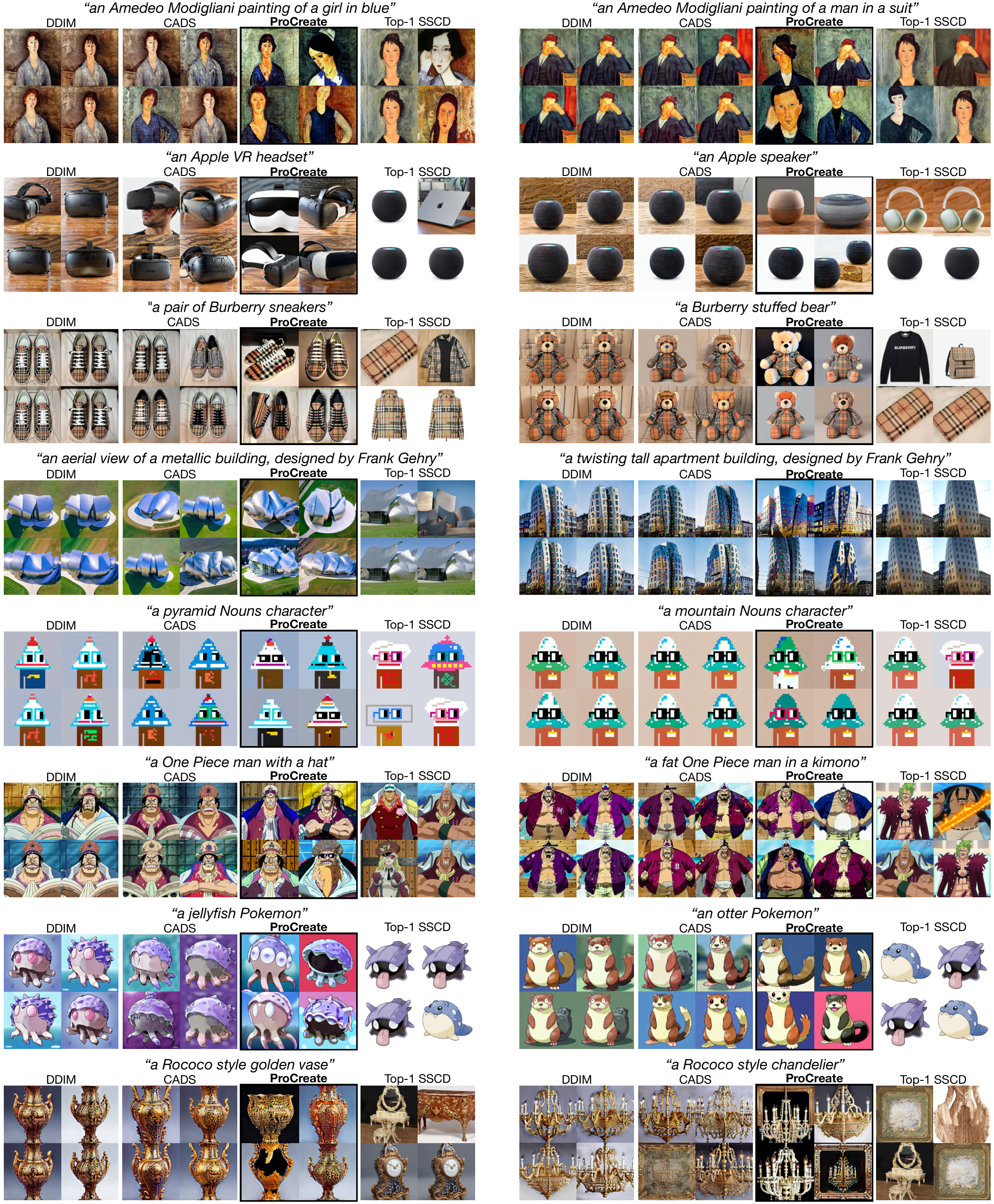}
  \caption{\looseness=-10000 Qualitative comparison between DDIM, CADS, and \ourmethod{} for few-shot creative generation on \ourdataset{} with standard fine-tuning. For each sampling method, we show two prompts and four generated samples for each prompt. In addition, we match each sample from \ourmethod{} with its closest training image based on the SSCD score \cite{sscd} between the matched pair.}
  \label{fig:normal_train10_qualitative}
  \vspace{-0.25in}
\end{figure}

\paragraph{Evaluation Metrics.}
\looseness=-10000 
We use FID \cite{heusel2017gans_fid} and KID \cite{binkowski2018demystifying} scores to measure how well our generated sample distribution matches that of real images, Precision and Recall \cite{improvedprecisionrecall} to measure the quality and diversity/coverage respectively, and following CADS, we use the Mean Similarity Score (MSS) and the Vendi score \cite{friedman2022vendi, sadat2023cads} to measure diversity of the generated samples. Specifically, we evaluate FID and KID with $64$ feature dimensions and set $k=5$ for Precision and Recall. To ensure that our generated images follow their input prompts, we also use CLIP to compute the cosine similarity between the prompt embedding and the generated image embedding, calling it Prompt Fidelity.

\subsection{Results}\label{subsec:fscg_results}

\paragraph{Creative Generation Visualization.}

We perform standard fine-tuning on a separate pre-trained stable diffusion for each dataset, then use the same checkpoint at 2000 iterations to perform DDIM, CADS, and \ourmethod{} sampling. Figure \ref{fig:normal_train10_qualitative} shows two example input prompts and their respective outputs from each sampling method. We observe that while the images produced by DDIM sampling do share similar properties and details as their 10-shot training images in Figure \ref{fig:datasets}, each image grid of 2-by-2 samples is very perceptually similar, lacking diversity. While CADS' samples can achieve better sample diversity than the default DDIM sampling, \ourmethod{} generates much more diverse images that balance between including shared properties in the training images and following the prompt guidance. Consider the generated samples of the prompt ``an Amedeo Modigliani painting of a girl in blue'', DDIM's $4$ samples are perceptually similar and only one of the CADS outputs significantly varies in posture and cloth color, while \ourmethod{} outputs contain women with diverse faces, hair color, background, and clothing. For the Top-1 SSCD \cite{sscd} columns, we select the closest training image to each \ourmethod{} sample based on the SSCD score, where a higher score suggests a higher likelihood of training data replication. Since all \ourmethod{} samples look significantly different from their matched training images, \ourmethod{} does not replicate its training images.

\begin{table}[h]
    \centering
    \caption{\looseness=-10000 Quantitative comparison between DDIM, CADS, and \ourmethod{} samples from standard fine-tuning checkpoints for $10$-shot learning on various generative modeling metrics. We show the mean and standard deviation values over 5 repeated runs in each cell.}
    \vspace{-0.1in}
    \resizebox{\textwidth}{!}{
        \begin{tabular}{|c|c|c|c|c|c|c|c|c|}
        \hline
            \bf Subset &  \bf Method &  \bf FID $\downarrow$ &  \bf KID $\downarrow$ &  \bf Precision $\uparrow$ &  \bf Recall $\uparrow$ &  \bf MSS $\downarrow$ &  \bf Vendi $\uparrow$ &  \bf Prompt Fid. $\uparrow$ \\ 
            \hline
            \multirow{3}{*}{Amedeo}   & DDIM      &     16.71 $\pm$ 0.39 &     2.39 $\pm$ 0.10 & \bf 0.75 $\pm$ 0.09 &     0.36 $\pm$ 0.06 &     0.34 $\pm$ 0.01 &     14.89 $\pm$ 0.69 & \bf 0.35 $\pm$ 0.00 \\
                                      & CADS      &     12.83 $\pm$ 0.48 &     2.29 $\pm$ 0.12 &     0.73 $\pm$ 0.05 &     0.39 $\pm$ 0.02 &     0.34 $\pm$ 0.01 &     15.17 $\pm$ 0.40 & \bf 0.35 $\pm$ 0.00 \\ 
                                      & \ourmethod{} & \bf  9.28 $\pm$ 0.39 & \bf 1.59 $\pm$ 0.25 &     0.57 $\pm$ 0.05 & \bf 0.65 $\pm$ 0.05 & \bf 0.26 $\pm$ 0.01 & \bf 22.20 $\pm$ 0.55 & \bf 0.35 $\pm$ 0.00 \\ 
            \hline
            \multirow{3}{*}{Apple}    & DDIM      &     48.51 $\pm$ 3.84 &     1.37 $\pm$ 0.13 & \bf 0.51 $\pm$ 0.03 &     0.75 $\pm$ 0.07 &     0.17 $\pm$ 0.01 &     25.13 $\pm$ 0.69 & \bf 0.30 $\pm$ 0.00 \\ 
                                      & CADS      &     38.88 $\pm$ 1.47 &     1.17 $\pm$ 0.22 &     0.43 $\pm$ 0.04 &     0.79 $\pm$ 0.01 &     0.17 $\pm$ 0.01 &     26.34 $\pm$ 0.87 & \bf 0.30 $\pm$ 0.00 \\ 
                                      & \ourmethod{} & \bf 24.59 $\pm$ 0.14 & \bf 0.62 $\pm$ 0.10 &     0.45 $\pm$ 0.06 & \bf 0.81 $\pm$ 0.02 & \bf 0.12 $\pm$ 0.00 & \bf 32.33 $\pm$ 0.37 &     0.29 $\pm$ 0.00 \\ 
            \hline
            \multirow{3}{*}{Burberry} & DDIM      &     35.11 $\pm$ 2.60 &     4.06 $\pm$ 0.48 & \bf 0.69 $\pm$ 0.06 &     0.71 $\pm$ 0.05 &     0.18 $\pm$ 0.00 &     26.15 $\pm$ 0.56 & \bf 0.34 $\pm$ 0.00 \\ 
                                      & CADS      &     37.71 $\pm$ 2.09 &     4.43 $\pm$ 0.37 & \bf 0.69 $\pm$ 0.07 &     0.74 $\pm$ 0.03 &     0.19 $\pm$ 0.01 &     25.46 $\pm$ 0.57 & \bf 0.34 $\pm$ 0.00 \\ 
                                      & \ourmethod{} & \bf 14.10 $\pm$ 1.64 & \bf 0.72 $\pm$ 0.11 &     0.66 $\pm$ 0.06 & \bf 0.97 $\pm$ 0.02 & \bf 0.10 $\pm$ 0.00 & \bf 33.51 $\pm$ 0.36 &     0.33 $\pm$ 0.00 \\ 
            \hline
            \multirow{3}{*}{Frank}    & DDIM      &      6.36 $\pm$ 0.37 &     0.37 $\pm$ 0.06 & \bf 0.99 $\pm$ 0.01 &     0.58 $\pm$ 0.05 &     0.17 $\pm$ 0.01 &     26.21 $\pm$ 0.57 & \bf 0.32 $\pm$ 0.00 \\ 
                                      & CADS      &      4.65 $\pm$ 0.32 &     0.35 $\pm$ 0.06 &    0.98 $\pm$ 0.02 &     0.59 $\pm$ 0.06 &     0.18 $\pm$ 0.01 &     26.49 $\pm$ 0.23 & \bf 0.32 $\pm$ 0.00 \\ 
                                      & \ourmethod{} & \bf  3.20 $\pm$ 0.24 & \bf 0.20 $\pm$ 0.02 &     0.94 $\pm$ 0.04 & \bf 0.67 $\pm$ 0.03 & \bf 0.16 $\pm$ 0.01 & \bf 29.70 $\pm$ 0.58 & \bf 0.32 $\pm$ 0.01 \\ 
            \hline
            \multirow{3}{*}{Nouns}    & DDIM      &      2.83 $\pm$ 0.04 &     0.04 $\pm$ 0.00 & \bf 0.12 $\pm$ 0.04 &     0.81 $\pm$ 0.11 &     0.47 $\pm$ 0.02 &     11.50 $\pm$ 0.51 & \bf 0.25 $\pm$ 0.00 \\ 
                                      & CADS      & \bf  2.54 $\pm$ 0.06 & \bf 0.03 $\pm$ 0.00 &     0.11 $\pm$ 0.03 &     0.82 $\pm$ 0.04 &     0.47 $\pm$ 0.01 &     11.50 $\pm$ 0.23 & \bf 0.25 $\pm$ 0.00 \\ 
                                      & \ourmethod{} &      2.72 $\pm$ 0.04 & \bf 0.03 $\pm$ 0.00 & \bf 0.12 $\pm$ 0.03 & \bf 0.91 $\pm$ 0.06 & \bf 0.42 $\pm$ 0.00 & \bf 12.07 $\pm$ 0.17 & \bf 0.25 $\pm$ 0.00 \\ 
            \hline
            \multirow{3}{*}{Onepiece} & DDIM      &      6.13 $\pm$ 0.50 &     0.67 $\pm$ 0.05 &     0.71 $\pm$ 0.04 &     0.64 $\pm$ 0.06 &     0.26 $\pm$ 0.01 &     23.73 $\pm$ 0.29 & \bf 0.30 $\pm$ 0.00 \\ 
            ~                         & CADS      &      7.13 $\pm$ 0.22 &     0.81 $\pm$ 0.11 &     0.71 $\pm$ 0.04 &     0.62 $\pm$ 0.02 &     0.27 $\pm$ 0.01 &     22.95 $\pm$ 0.40 & \bf 0.30 $\pm$ 0.00 \\ 
            ~                         & \ourmethod{} & \bf  4.84 $\pm$ 0.24 & \bf 0.44 $\pm$ 0.07 & \bf 0.72 $\pm$ 0.05 & \bf 0.67 $\pm$ 0.01 & \bf 0.25 $\pm$ 0.01 & \bf 25.80 $\pm$ 0.27 & \bf 0.30 $\pm$ 0.00 \\ 
            \hline
            \multirow{3}{*}{Pokemon}  & DDIM      &     14.29 $\pm$ 0.94 &     0.28 $\pm$ 0.02 &     0.44 $\pm$ 0.07 &     0.77 $\pm$ 0.10 &     0.30 $\pm$ 0.01 &     21.46 $\pm$ 0.60 & \bf 0.32 $\pm$ 0.00 \\ 
            ~                         & CADS      &     16.57 $\pm$ 0.39 &     0.28 $\pm$ 0.02 & \bf 0.48 $\pm$ 0.10 &     0.75 $\pm$ 0.03 &     0.29 $\pm$ 0.01 &     22.29 $\pm$ 0.35 & \bf 0.32 $\pm$ 0.00 \\ 
            ~                         & \ourmethod{} & \bf 11.38 $\pm$ 0.76 & \bf 0.27 $\pm$ 0.01 &     0.44 $\pm$ 0.07 & \bf 0.84 $\pm$ 0.04 & \bf 0.28 $\pm$ 0.01 & \bf 22.51 $\pm$ 0.37 &     0.31 $\pm$ 0.00 \\ 
            \hline
            \multirow{3}{*}{Rococo}   & DDIM      &     26.47 $\pm$ 1.17 &     5.64 $\pm$ 0.41 & \bf 0.95 $\pm$ 0.01 &     0.83 $\pm$ 0.03 &     0.16 $\pm$ 0.00 &     22.68 $\pm$ 0.28 & \bf 0.33 $\pm$ 0.00 \\ 
                                      & CADS      &     30.10 $\pm$ 1.35 &     6.63 $\pm$ 0.43 &     0.92 $\pm$ 0.02 &     0.78 $\pm$ 0.04 &     0.17 $\pm$ 0.00 &     22.22 $\pm$ 0.28 & \bf 0.33 $\pm$ 0.00 \\ 
                                      & \ourmethod{} & \bf 17.96 $\pm$ 1.44 & \bf 3.26 $\pm$ 0.36 & \bf 0.95 $\pm$ 0.03 & \bf 0.89 $\pm$ 0.02 & \bf 0.10 $\pm$ 0.01 & \bf 23.43 $\pm$ 1.15 &     0.32 $\pm$ 0.00 \\ 
            \hline
        \end{tabular}
    }
    \label{tab:normal_train10}
\end{table}

\begin{table}[h]
    \centering
    \caption{\looseness=-10000 Quantitative comparison between DDIM, CADS, and \ourmethod{} samples from DreamBooth fine-tuning checkpoints for $10$-shot learning on various generative modeling metrics. We show the mean and standard deviation values over 5 repeated runs in each cell.}
    \vspace{-0.1in}
    \resizebox{\textwidth}{!}{
        \begin{tabular}{|c|c|c|c|c|c|c|c|c|}
        \hline
            \bf Subset &  \bf Method &  \bf FID $\downarrow$ &  \bf KID $\downarrow$ &  \bf Precision $\uparrow$ &  \bf Recall $\uparrow$ &  \bf MSS $\downarrow$ &  \bf Vendi $\uparrow$ &  \bf Prompt Fid. $\uparrow$ \\ 
            \hline
            \multirow{3}{*}{Amedeo}   & DDIM      &     17.38 $\pm$ 0.40 &     2.96 $\pm$ 0.13 &     0.48 $\pm$ 0.04 &     0.63 $\pm$ 0.15 &     0.28 $\pm$ 0.00 &     14.12 $\pm$ 0.37 & \bf 0.32 $\pm$ 0.00 \\
                                      & CADS      &     16.19 $\pm$ 0.56 &     2.56 $\pm$ 0.17 & \bf 0.53 $\pm$ 0.06 &     0.65 $\pm$ 0.07 &     0.27 $\pm$ 0.00 &     14.37 $\pm$ 0.40 & \bf 0.32 $\pm$ 0.00 \\ 
                                      & \ourmethod{} & \bf  8.88 $\pm$ 0.17 & \bf 1.19 $\pm$ 0.27 &     0.45 $\pm$ 0.05 & \bf 0.80 $\pm$ 0.01 & \bf 0.20 $\pm$ 0.01 & \bf 25.49 $\pm$ 0.55 &     0.31 $\pm$ 0.00 \\ 
            \hline
            \multirow{3}{*}{Apple}    & DDIM      &     29.71 $\pm$ 0.63 &     0.70 $\pm$ 0.07 &     0.39 $\pm$ 0.02 &     0.71 $\pm$ 0.05 &     0.22 $\pm$ 0.01 &     17.38 $\pm$ 1.04 & \bf 0.27 $\pm$ 0.00 \\ 
                                      & CADS      &     32.52 $\pm$ 2.21 &     0.84 $\pm$ 0.12 &     0.35 $\pm$ 0.06 &     0.71 $\pm$ 0.02 &     0.21 $\pm$ 0.01 &     18.82 $\pm$ 1.39 & \bf 0.27 $\pm$ 0.00 \\ 
                                      & \ourmethod{} & \bf 20.12 $\pm$ 2.44 & \bf 0.27 $\pm$ 0.05 & \bf 0.40 $\pm$ 0.03 & \bf 0.84 $\pm$ 0.04 & \bf 0.15 $\pm$ 0.00 & \bf 28.23 $\pm$ 0.83 & \bf 0.27 $\pm$ 0.00 \\ 
            \hline
            \multirow{3}{*}{Burberry} & DDIM      &     21.89 $\pm$ 2.24 &     2.28 $\pm$ 0.33 &     0.56 $\pm$ 0.06 &     0.78 $\pm$ 0.07 &     0.20 $\pm$ 0.01 &     20.85 $\pm$ 0.57 & \bf 0.28 $\pm$ 0.00 \\ 
                                      & CADS      &     19.99 $\pm$ 1.53 &     2.10 $\pm$ 0.36 &     0.54 $\pm$ 0.07 &     0.81 $\pm$ 0.12 &     0.21 $\pm$ 0.00 &     20.26 $\pm$ 0.40 &     0.27 $\pm$ 0.00 \\ 
                                      & \ourmethod{} & \bf 10.18 $\pm$ 1.03 & \bf 0.37 $\pm$ 0.06 & \bf 0.59 $\pm$ 0.06 & \bf 0.92 $\pm$ 0.05 & \bf 0.12 $\pm$ 0.00 & \bf 28.64 $\pm$ 0.85 & \bf 0.28 $\pm$ 0.00 \\ 
            \hline
            \multirow{3}{*}{Frank}    & DDIM      &      3.88 $\pm$ 0.14 &     0.34 $\pm$ 0.02 & \bf 0.71 $\pm$ 0.02 &     0.65 $\pm$ 0.01 &     0.20 $\pm$ 0.00 &     18.66 $\pm$ 0.30 & \bf 0.26 $\pm$ 0.00 \\ 
                                      & CADS      &      3.42 $\pm$ 0.27 &     0.31 $\pm$ 0.05 &     0.70 $\pm$ 0.02 &     0.70 $\pm$ 0.02 &     0.20 $\pm$ 0.00 &     18.62 $\pm$ 0.28 & \bf 0.26 $\pm$ 0.00 \\ 
                                      & \ourmethod{} & \bf  2.38 $\pm$ 0.18 & \bf 0.09 $\pm$ 0.01 &     0.57 $\pm$ 0.03 & \bf 0.78 $\pm$ 0.02 & \bf 0.14 $\pm$ 0.01 & \bf 26.99 $\pm$ 1.41 & \bf 0.26 $\pm$ 0.00 \\ 
            \hline
            \multirow{3}{*}{Nouns}    & DDIM      &      0.65 $\pm$ 0.02 &     0.02 $\pm$ 0.00 & \bf 0.77 $\pm$ 0.07 &     0.39 $\pm$ 0.03 &     0.61 $\pm$ 0.01 &      6.50 $\pm$ 0.17 & \bf 0.25 $\pm$ 0.00 \\ 
                                      & CADS      &      0.66 $\pm$ 0.02 &     0.04 $\pm$ 0.00 &     0.76 $\pm$ 0.03 &     0.38 $\pm$ 0.03 &     0.61 $\pm$ 0.00 &      6.41 $\pm$ 0.09 & \bf 0.25 $\pm$ 0.00 \\ 
                                      & \ourmethod{} & \bf  0.62 $\pm$ 0.01 & \bf 0.01 $\pm$ 0.00 &     0.76 $\pm$ 0.04 & \bf 0.43 $\pm$ 0.02 & \bf 0.56 $\pm$ 0.01 & \bf  7.24 $\pm$ 0.18 & \bf 0.25 $\pm$ 0.00 \\ 
            \hline
            \multirow{3}{*}{Onepiece} & DDIM      &      5.16 $\pm$ 0.26 &     0.32 $\pm$ 0.03 & \bf 0.36 $\pm$ 0.01 &     0.71 $\pm$ 0.02 &     0.23 $\pm$ 0.00 &     18.72 $\pm$ 0.19 & \bf 0.25 $\pm$ 0.00 \\ 
            ~                         & CADS      &      5.33 $\pm$ 0.30 &     0.36 $\pm$ 0.04 &     0.26 $\pm$ 0.02 &     0.70 $\pm$ 0.02 &     0.23 $\pm$ 0.00 &     18.78 $\pm$ 0.41 & \bf 0.25 $\pm$ 0.00 \\ 
            ~                         & \ourmethod{} & \bf  4.34 $\pm$ 0.19 & \bf 0.13 $\pm$ 0.03 &     0.28 $\pm$ 0.02 & \bf 0.86 $\pm$ 0.02 & \bf 0.20 $\pm$ 0.00 & \bf 19.98 $\pm$ 0.38 & \bf 0.25 $\pm$ 0.00 \\ 
            \hline
            \multirow{3}{*}{Pokemon}  & DDIM      & \bf  8.13 $\pm$ 0.82 & \bf 0.13 $\pm$ 0.03 & \bf 0.46 $\pm$ 0.04 &     0.85 $\pm$ 0.03 &     0.25 $\pm$ 0.00 &     21.08 $\pm$ 0.68 & \bf 0.25 $\pm$ 0.00 \\ 
            ~                         & CADS      &     13.51 $\pm$ 0.89 &     0.25 $\pm$ 0.03 &     0.38 $\pm$ 0.03 &     0.88 $\pm$ 0.05 &     0.21 $\pm$ 0.01 &     24.32 $\pm$ 0.75 & \bf 0.25 $\pm$ 0.00 \\ 
            ~                         & \ourmethod{} &     11.67 $\pm$ 0.47 &     0.20 $\pm$ 0.03 &     0.41 $\pm$ 0.06 & \bf 0.92 $\pm$ 0.04 & \bf 0.17 $\pm$ 0.02 & \bf 27.70 $\pm$ 1.77 & \bf 0.25 $\pm$ 0.00 \\ 
            \hline
            \multirow{3}{*}{Rococo}   & DDIM      &     13.57 $\pm$ 1.09 &     2.00 $\pm$ 0.28 & \bf 0.93 $\pm$ 0.00 &     0.77 $\pm$ 0.02 &     0.17 $\pm$ 0.00 &     12.10 $\pm$ 0.96 &     0.24 $\pm$ 0.00 \\ 
                                      & CADS      &     16.27 $\pm$ 1.26 &     2.59 $\pm$ 0.30 &     0.83 $\pm$ 0.00 &     0.80 $\pm$ 0.01 &     0.17 $\pm$ 0.00 &     12.15 $\pm$ 0.40 &     0.24 $\pm$ 0.00 \\ 
                                      & \ourmethod{} & \bf  9.35 $\pm$ 1.06 & \bf 1.12 $\pm$ 0.19 &     0.88 $\pm$ 0.05 & \bf 0.85 $\pm$ 0.02 & \bf 0.12 $\pm$ 0.01 & \bf 23.07 $\pm$ 1.95 & \bf 0.25 $\pm$ 0.00 \\ 
            \hline
        \end{tabular}
    }
    \label{tab:dreambooth_train10}
    \vspace{-0.1in}
\end{table}

\paragraph{Quantitative Evaluation with Standard Fine-Tuning.} In Table \ref{tab:normal_train10}, we use the metrics in Section \ref{sec:fscg_experiment_setup} 
to evaluate the same set of standard fine-tuning checkpoints used for qualitative results. Due to the small dataset size, we evaluate each checkpoint and method $5$ times with random train/validation splits to obtain the mean and standard deviation values for each metric. On close to all datasets of \ourdataset{}, not only does \ourmethod{} achieve significantly better performance than DDIM and CADS in all diversity-focused metrics (Recall, MSS, Vendi), \ourmethod{} is also superior in metrics that measure both quality and diversity (FID, KID). \ourmethod{} being competitive with other sampling methods in Precision and Prompt Fidelity shows that it can generate diverse samples while preserving high output quality and fidelity.

\begin{figure}[h]
  \centering
  \includegraphics[width=\textwidth,center]{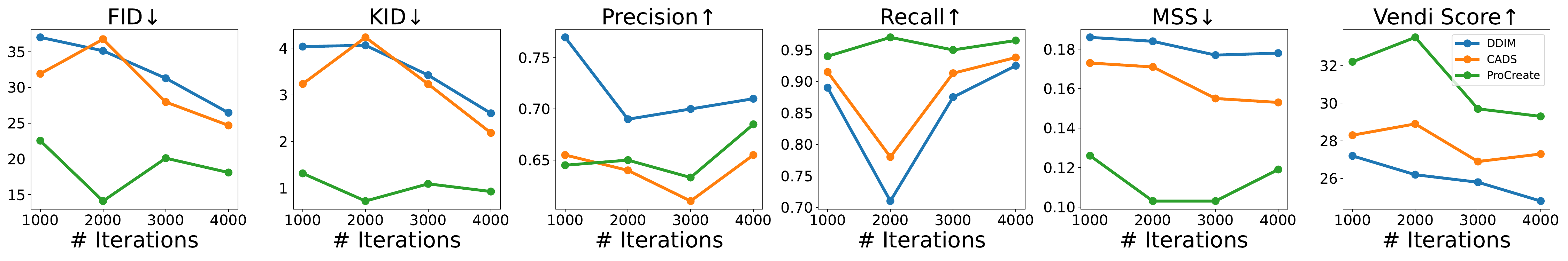}
  \caption{Compare DDIM, CADS, and \ourmethod{}'s performance under different numbers of fine-tuning iterations.}
  \label{fig:ckpt_ablation}
  \vspace{-0.2in}
\end{figure}

\begin{figure}[h]
  \centering
  \includegraphics[width=\textwidth,center]{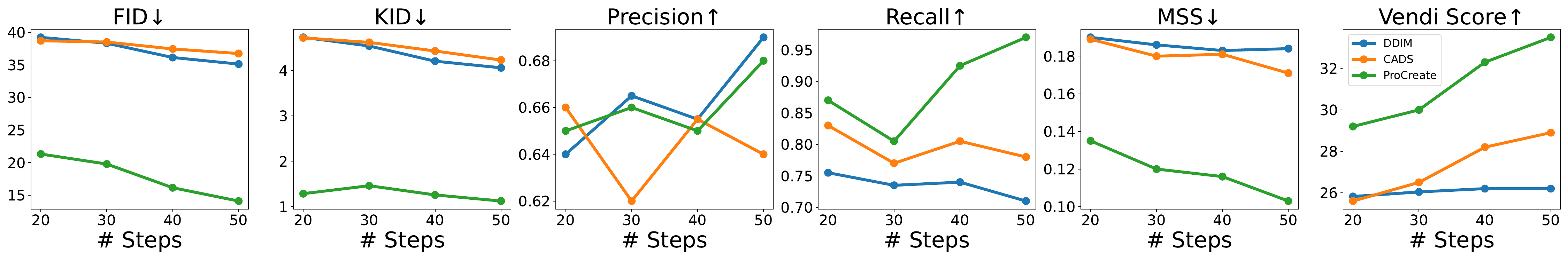}
  \caption{Compare DDIM, CADS, and \ourmethod{}'s performance under different numbers of inference steps.}
  \label{fig:nstep_ablation}
  \vspace{-0.2in}
\end{figure}

\paragraph{Quantitative Comparison with DreamBooth Fine-Tuning.}
\looseness=-10000
In addition to standard fine-tuning, we can also apply \ourmethod{} on DreamBooth~\cite{ruiz2023dreambooth}, where a DreamBooth model is fine-tuned on each category of the few-shot dataset. Table \ref{tab:dreambooth_train10} shows the evaluation results for checkpoints obtained from DreamBooth training. \ourmethod{} again achieves significantly better performance on FID, KID, Recall, MSS, and Vendi Score on almost all datasets while remaining competitive with our baselines in Precision and Prompt Fidelity. The results of this experiment show that \ourmethod{} improves the performance of a state-of-the-art few-shot learning method and gives strong evidence for \ourmethod{}'s general ability to improve any checkpoint fine-tuned on limited samples. 

\subsection{Ablation Experiments}\label{subsec:ablation_experiments}

\paragraph{Fine-Tuning Steps.}
\looseness=-10000
We evaluate the baselines and \ourmethod{} on checkpoints at training iterations 1k, 2k, 3k, and 4k of the standard fine-tuning run on the ``Burberry Designs'' dataset. Figure \ref{fig:ckpt_ablation} shows that \ourmethod{} consistently improves FID, KID, Recall, MSS, and Vendi Score and improves the trade-off in Precision in comparison to CADS. 

\paragraph{Number of Inference Steps.} 
\looseness=-10000
We evaluate the baselines and \ourmethod{} on the number of diffusion inference steps from 20 to 50. \ourmethod{} again shows consistent improvements in all metrics except Precision while achieving a better trade-off in Precision in comparison to CADS.

\paragraph{Additional Ablation Results.}
\looseness=-10000
We also experimented with varying the diffusion samplers (e.g., DDPM, PNDM \cite{liu2022pseudo}), the number of training samples (e.g., 5-shot learning, 25-shot learning), and varying the number of steps for Multi-Step Look Ahead prediction. These results are included in the Supplementary Materials.

\section{Training Data Replication Prevention}\label{sec:replication}

\looseness=-10000
With the surging interest in generative AI in recent years, people use image generation models for a variety of entertainment or business purposes. However, recent studies show that even large-scale diffusion models are prone to replicating data inside its training set \cite{somepalli2023diffusion_forgery, somepalli2023diffusion_copying}, raising privacy and copyright concerns. To prevent large-scale diffusion models from replicating their training data, we follow the setup of Somepalli \etal \cite{somepalli2023diffusion_forgery} for this experiment to test \ourmethod{}'s ability to guide generations of pre-trained Stable Diffusion models away from their LAION \cite{LAION} training dataset.

\begin{figure}[!h]
\vspace{-0.15in}
  \centering
  \includegraphics[width=0.95\textwidth,trim={0cm 3.5cm 0cm 0.7cm},clip]{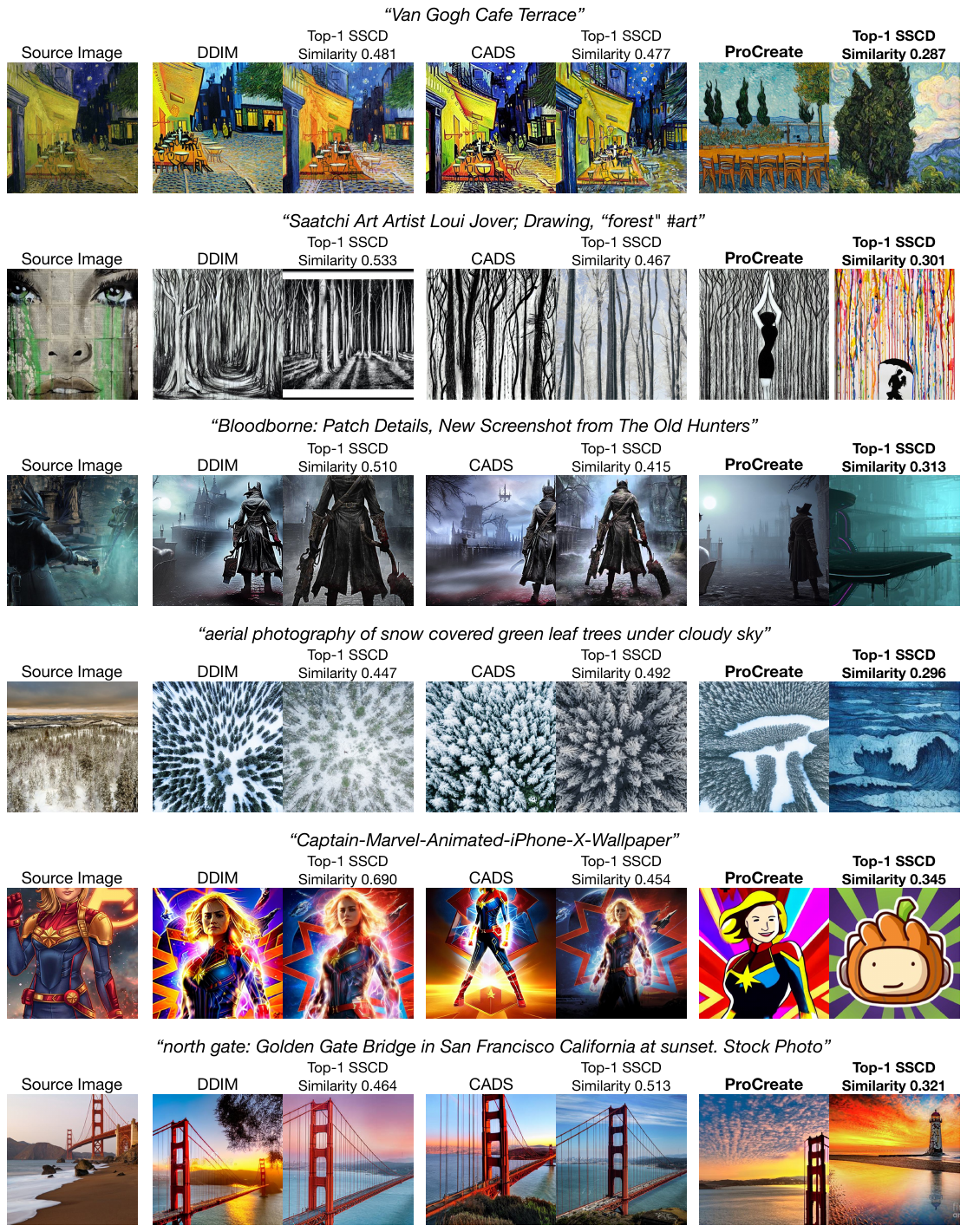}
  \caption{Qualitative comparison between DDIM, CADS, and \ourmethod{} for replication prevention on LAION12M.}
  \label{fig:replication_qualitative}
  \vspace{-0.2in}
\end{figure}

\subsection{Experiment Setup}\label{replication_experiment_setup}

\paragraph{Model and Dataset.} 
\looseness=-10000
We use the frozen pre-trained Stable Diffusion v1-5 checkpoint that is pre-trained on the LAION dataset containing 2B prompt-image pairs. To scale the compute within our limitation, we follow Somepalli \etal \cite{somepalli2023diffusion_forgery} by setting the reference set to the smaller LAION Aesthetics v2 6+ dataset with 12M caption-image pairs and is a subset of images used in the last stage of Stable Diffusion v1-5 fine-tuning, namely LAION12M. Using a random subset of prompts from LAION12M, we generate 9k samples with Stable Diffusion v1-5 and search each sample's matching LAION12M image with the highest Top-1 SSCD score.

\paragraph{Inference Implementation.}
\looseness=-10000
We perform 50 inference steps and set \ourmethod{}'s Multi-Step Look Ahead $n_{step}$ to $5$. To perform inference efficiently, before generating each sample, we filter LAION12M to 10k images that have the most similar captions to the sample caption based on their CLIP embeddings' cosine similarity. We use the FAISS \cite{faiss} library to speed up vector similarity searches. We again compare \ourmethod{} to DDIM and CADS.

\paragraph{Evaluation Metrics.}
\looseness=-10000
To evaluate how well \ourmethod{} prevents data replication, we compute the percentages of Top-1 SSCD scores over the thresholds $0.4$, $0.5$, and $0.6$. To ensure that the generated images are still within the distribution of LAION12M images, we also compare them with FID and KID.

\subsection{Results}\label{subsec:results}

\paragraph{Image Generation Visualization.} 
\looseness=-10000
Figure \ref{fig:replication_qualitative} shows examples of when the pre-trained Stable Diffusion model generates images that are both perceptually similar to their matched images in LAION12M and high in Top-1 SSCD score. While CADS sampling reduces the Top-1 SSCD score in most cases, the ``Van Gogh Cafe Terrace'' and ``Golden Bridge'' examples show that CADS is insufficient for preventing replication. In contrast, \ourmethod{} significantly lowers the perceptual similarity and Top-1 SSCD scores in all examples, showing that their generated samples are sufficiently different in SSCD score ($<0.4$) from all other images in LAION12M. This can be explained by \ourmethod{}'s ability to dynamically select the closest LAION12M example to be propelled from during guidance so that the generated image would always be guided away from an arbitrary LAION12M image that it is close to. Since DreamSim imitates human perception for detecting similarities in images, using it as our similarity embedding network ensures that \ourmethod{} outputs do not replicate training images.

\begin{figure}[t]
  \begin{minipage}{.48\linewidth}
    \centering
    \includegraphics[width=1.0\linewidth]{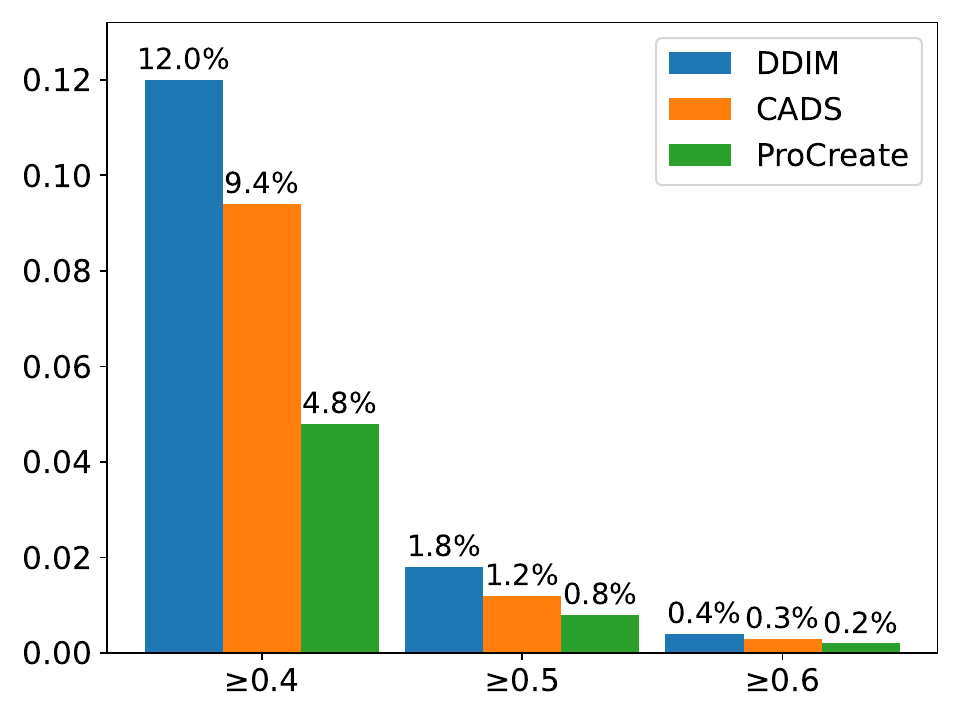}
    \captionof{figure}{Comparison of DDIM, CADS, and \ourmethod{} on the percentage of generated 9k images over a Top-1 SSCD threshold.}
    \label{fig:top1_barplot}
  \end{minipage}
  \hfill
  \begin{minipage}{.48\linewidth}
    \begin{minipage}[c]{1\linewidth}
        \centering
        \begin{tabular}{ |c|c|c| } 
             \hline
             & \textbf{FID} & \textbf{KID} \\ 
             \hline
             \textbf{DDIM} & 1.38 & 0.051 \\
             \hline
             \textbf{CADS} & 1.60 & 0.065 \\
             \hline
             \textbf{\ourmethod{}} & \bf 1.14 & \bf 0.038 \\ 
             \hline
        \end{tabular}
        \captionof{table}{\looseness=-10000 Comparison of DDIM, CADS, and \ourmethod{} on FID and KID.}
        \label{tab:replication_fid_kid}
    \end{minipage}

    \vspace{0.1in}
    
    \begin{minipage}{1\linewidth}
        \centering
        \begin{tabular}{|c|c|}
            \hline
            $\bm{n_{step}}$ &  \bf Inference Time (s/sample) \\
            \hline
            0 &     3.1 \\
            1 &     18.5 \\
            3 &     28.3 \\
            5 &     37.0 \\
            \hline
        \end{tabular}
        \vspace{-0.1in}
        \captionof{table}{\looseness=-10000 MSLA $n_{step}$'s effect on \ourmethod{}'s inference time with NVIDIA A100.}
        \label{tab:inference_time}
    \end{minipage}
  \end{minipage}
  \vspace{-0.2in}
\end{figure}
    
\paragraph{Top-1 SSCD Scores.}
\looseness=-10000
In Figure \ref{fig:top1_barplot}, we compare the percentage of top1-SSCD scores of each 10k images generated with DDIM, CADS, and \ourmethod{} sampling. \ourmethod{} reduces the percentages of DDIM generations by more than half, demonstrating superior ability in preventing data replication from the pre-trained Stable Diffusion model. 

\paragraph{Comparison on Distribution Metrics.}
\looseness=-10000
Interestingly, Table \ref{tab:replication_fid_kid} shows that \ourmethod{} also reduced both FID and KID when compared to the baseline DDIM sampling while CADS does not, indicating that \ourmethod{} not only reduces data replication but also improves the generation quality with higher sample diversity.

\section{Conclusion and Discussion}\label{conclusion_and_discussion}

\begin{figure}[t]
  \centering
  \includegraphics[width=0.95\textwidth,trim={1.3cm 20cm 1.3cm 0.7cm},clip]{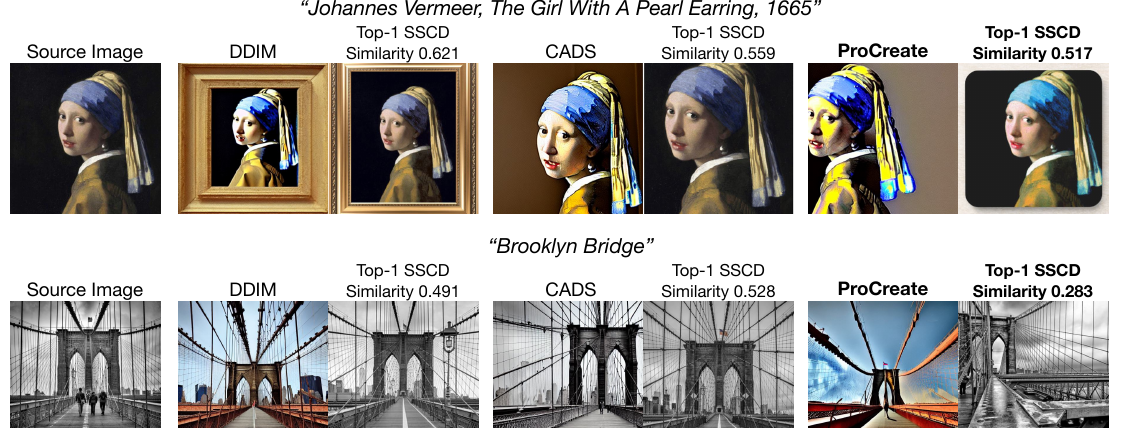}
  \caption{Failure cases from the LAION12M experiment in Section \ref{sec:replication}.}
  \label{fig:laion12m_replication_failure}
  \vspace{-0.25in}  
\end{figure}

\looseness=-10000
We propose \ourmethod{}, a simple method for improving the generation diversity and creativity of diffusion models given a set of reference images. We focus on two setups to demonstrate the effectiveness of \ourmethod{}. First, in the low-data fine-tuning regime, we introduce a new text-to-image few-shot generation dataset \ourdataset{} and we show that \ourmethod{} achieves the best performance in terms of sample diversity and distribution metrics compared to prior approaches. Second, for pre-trained diffusion networks, we show that \ourmethod{} can effectively mitigate training data replication and improve sample quality on the LAION dataset. In the future, we expect that \ourmethod{} can be applied to more modalities of data, including audio and video, to promote more diverse and creative generation of digital content. 

\paragraph{Broader Impact.} 
\looseness=-10000
This research presents significant broader impacts. It offers content creators and designers the tools to enhance AI-assisted design with a smaller risk of replicating reference images, or private/copyrighted training images. Although the primary implications are beneficial, there exists a potential for this technology to facilitate the design of counterfeit products. Addressing the ethical use and regulatory oversight of such advancements warrants further discussion in future works.

\paragraph{Limitations.}
\looseness=-10000
Currently, there are several limitations to \ourmethod{}. The guidance process is slow compared to direct generation since it performs Multi-Step Look Ahead and needs to backpropagate through a similarity network. In Table \ref{tab:inference_time}, we show \ourmethod{}'s generation time at different $n_{step}$ values with batch size 1 and 50 inference steps on an NVIDIA A100 GPU. The method can also sometimes be sensitive to the guidance strength parameter $\gamma$, requiring extra hyperparameter tuning on new datasets. Figure \ref{fig:laion12m_replication_failure} shows two failure scenarios where \ourmethod{} sample replicates the Top-1 SSCD matched training image in the first row due to low guidance strength, and where \ourmethod{}'s sample quality degrades when the guidance strength is too high in the second row.

\section*{Acknowledgement}
\looseness=-10000
We thank Zhun Deng and members of the NYU Agentic Learning AI Lab for their helpful discussions. The NYU High-Performance Computing resources, services, and staff expertise supported the compute.

\bibliographystyle{splncs04}
\bibliography{main}

\clearpage

\section{Additional Results}
\begin{figure}[h!]
  \centering
  \includegraphics[width=0.8\textwidth]{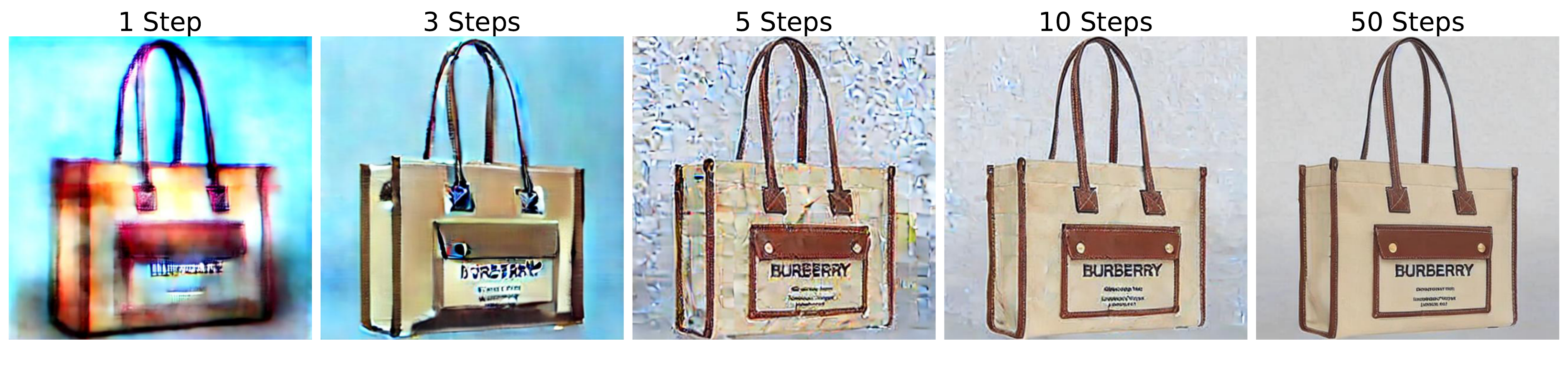}
  \vspace{-0.1in}
  \caption{Predictions for ``Burberry Tote Bag'' using different numbers of DDIM sampling steps from the evaluated Burberry checkpoint in Table \ref{tab:normal_train10}.}
  \label{fig:nstep_ablation_qualitative}
\end{figure}

\paragraph{Varying Multi-Step Look Ahead $n_{step}$.}
\looseness=-10000
Section \ref{sec:background} shows that at each denoising step, $\hat x_0$ can be predicted in a single step by Equation \ref{eq:onestep}. However, the results displayed in Figure \ref{fig:nstep_ablation_qualitative} indicate that the 1-step prediction may be blurry and inaccurate. As more DDIM sampling steps are taken, the image quality increases, leading to an improvement in the similarity embedding and the propulsive guidance gradient. Table \ref{tab:nstep_ablation_quantitative} further illustrates that while \ourmethod{} still outperforms DDIM sampling when the Multi-Step Look Ahead method is not used ($n_{step}=1$), the method significantly improves FID, KID, and diversity-focused metrics at higher $n_{step}$ settings.

\vspace{-0.3in}

\begin{table}[h!]
    \centering
    \caption{The effect of Multi-Step Look Ahead $n_{step}$ on \ourmethod{}'s Performance. $n_{step}=0$ represents baseline DDIM sampling that does not use \ourmethod{}.}
    \vspace{-0.1in}
    \resizebox{\textwidth}{!}{
        \begin{tabular}{|c|c|c|c|c|c|c|c|}
        \hline
            \bf MSLA $n_{step}$ &  \bf FID $\downarrow$ &  \bf KID $\downarrow$ &  \bf Precision $\uparrow$ &  \bf Recall $\uparrow$ &  \bf MSS $\downarrow$ &  \bf Vendi $\uparrow$ &  \bf Prompt Fid. $\uparrow$ \\ 
            \hline
            0 &     35.11 $\pm$ 2.60 &     4.06 $\pm$ 0.48 & \bf 0.69 $\pm$ 0.06 &     0.71 $\pm$ 0.05 &     0.18 $\pm$ 0.00 &     26.15 $\pm$ 0.56 & \bf 0.34 $\pm$ 0.00 \\
            1 &     29.99 $\pm$ 1.49 &     2.64 $\pm$ 0.08 &     0.60 $\pm$ 0.06 &     0.91 $\pm$ 0.05 &     0.13 $\pm$ 0.01 &     30.90 $\pm$ 0.69 &     0.33 $\pm$ 0.00 \\ 
            3 &     17.82 $\pm$ 2.99 &     0.79 $\pm$ 0.47 &     0.64 $\pm$ 0.02 &     0.91 $\pm$ 0.04 &     0.10 $\pm$ 0.01 &     33.51 $\pm$ 0.58 &     0.33 $\pm$ 0.00 \\
            5 & \bf 14.10 $\pm$ 1.64 & \bf 0.72 $\pm$ 0.11 &     0.66 $\pm$ 0.06 & \bf 0.97 $\pm$ 0.02 & \bf 0.09 $\pm$ 0.00 & \bf 34.12 $\pm$ 0.36 &     0.33 $\pm$ 0.00 \\
            \hline
        \end{tabular}
    }
    \label{tab:nstep_ablation_quantitative}
    \vspace{-0.1in}
\end{table}

\paragraph{Using Different Base Diffusion Samplers.}
\looseness=-10000
We experiment with different base diffusion samplers using the standard fine-tuned "Burberry Designs'' checkpoint from Table \ref{tab:normal_train10}. Table \ref{tab:sampler_ablation} shows that although the DDPM and PNDM base samplers does not perform as well as the DDIM sampler on generative modeling metrics, we can apply \ourmethod{} to them to improve their output diversity while maintaining sample quality.

\begin{table}[h]
    \centering
    \caption{\looseness=-10000 Quantitative comparison between DDIM, CADS, and \ourmethod{} on different base samplers DDIM, DDPM, and PNDM. The evaluation is performed with standard fine-tuning, ``Burberry Designs'' images, and a $10$ training and $40$ validation split.}
    \resizebox{\textwidth}{!}{
        \begin{tabular}{|c|c|c|c|c|c|c|c|c|}
        \hline
            \bf Sampler &  \bf Method &  \bf FID $\downarrow$ &  \bf KID $\downarrow$ &  \bf Precision $\uparrow$ &  \bf Recall $\uparrow$ &  \bf MSS $\downarrow$ &  \bf Vendi $\uparrow$ &  \bf Prompt Fid. $\uparrow$ \\ 
            \hline
            \multirow{3}{*}{DDIM}   & Default   &     35.11 $\pm$ 2.60 &     4.06 $\pm$ 0.48 &     0.69 $\pm$ 0.06 &     0.71 $\pm$ 0.05 &     0.18 $\pm$ 0.00 &     26.15 $\pm$ 0.56 & \bf 0.34 $\pm$ 0.00 \\
                                    & CADS      &     34.39 $\pm$ 1.97 &     4.27 $\pm$ 0.41 & \bf 0.70 $\pm$ 0.07 &     0.81 $\pm$ 0.06 &     0.18 $\pm$ 0.00 &     25.97 $\pm$ 0.20 & \bf 0.34 $\pm$ 0.00 \\ 
                                    & \ourmethod{} & \bf 14.10 $\pm$ 1.64 & \bf 0.72 $\pm$ 0.11 &     0.66 $\pm$ 0.06 & \bf 0.97 $\pm$ 0.02 & \bf 0.10 $\pm$ 0.00 & \bf 33.51 $\pm$ 0.36 &     0.33 $\pm$ 0.00 \\ 
            \hline
            \multirow{3}{*}{DDPM}   & Default   &     37.29 $\pm$ 1.97 &     4.08 $\pm$ 0.27 & \bf 0.68 $\pm$ 0.05 &     0.70 $\pm$ 0.07 &     0.19 $\pm$ 0.01 &     26.05 $\pm$ 0.68 & \bf 0.34 $\pm$ 0.00 \\
                                    & CADS      &     37.24 $\pm$ 1.81 &     4.36 $\pm$ 0.31 & \bf 0.68 $\pm$ 0.03 &     0.73 $\pm$ 0.08 &     0.19 $\pm$ 0.01 &     25.62 $\pm$ 0.48 & \bf 0.34 $\pm$ 0.00 \\ 
                                    & \ourmethod{} & \bf 14.80 $\pm$ 2.18 & \bf 2.03 $\pm$ 0.22 &     0.67 $\pm$ 0.08 & \bf 0.95 $\pm$ 0.03 & \bf 0.12 $\pm$ 0.01 & \bf 31.74 $\pm$ 0.62 & \bf 0.34 $\pm$ 0.00 \\ 
            \hline
            \multirow{3}{*}{PNDM}   & Default   &     60.86 $\pm$ 2.22 &     8.94 $\pm$ 0.53 & \bf 0.64 $\pm$ 0.09 &     0.49 $\pm$ 0.10 &     0.22 $\pm$ 0.01 &     24.51 $\pm$ 1.56 & \bf 0.30 $\pm$ 0.01 \\
                                    & CADS      &     63.24 $\pm$ 1.47 &     9.44 $\pm$ 0.42 & \bf 0.64 $\pm$ 0.04 &     0.56 $\pm$ 0.06 &     0.23 $\pm$ 0.01 &     23.50 $\pm$ 0.54 & \bf 0.30 $\pm$ 0.00 \\ 
                                    & \ourmethod{} & \bf 28.64 $\pm$ 2.09 & \bf 2.22 $\pm$ 0.34 &     0.62 $\pm$ 0.05 & \bf 0.92 $\pm$ 0.04 & \bf 0.11 $\pm$ 0.01 & \bf 32.08 $\pm$ 0.44 & \bf 0.30 $\pm$ 0.00 \\ 
            \hline
        \end{tabular}
    }
    \label{tab:sampler_ablation}
    \vspace{-0.1in}
\end{table}

\begin{table}
    \centering
    \caption{\looseness=-10000 Quantitative comparison between DDIM, CADS, and \ourmethod{} applied on standard fine-tuning checkpoints for $5$-shot learning on various generative modeling metrics. We show the mean and standard deviation values over 5 repeated runs in each cell.}
    \resizebox{\textwidth}{!}{
        \begin{tabular}{|c|c|c|c|c|c|c|c|c|}
        \hline
            \bf Subset &  \bf Method &  \bf FID $\downarrow$ &  \bf KID $\downarrow$ &  \bf Precision $\uparrow$ &  \bf Recall $\uparrow$ &  \bf MSS $\downarrow$ &  \bf Vendi $\uparrow$ &  \bf Prompt Fid. $\uparrow$ \\ 
            \hline
            \multirow{3}{*}{Amedeo}   & DDIM      &     11.27 $\pm$ 0.69 &     2.02 $\pm$ 0.18 & \bf 0.81 $\pm$ 0.03 &     0.44 $\pm$ 0.12 &     0.36 $\pm$ 0.01 &     14.32 $\pm$ 0.70 & \bf 0.34 $\pm$ 0.00 \\
                                      & CADS      &     12.51 $\pm$ 0.68 &     2.43 $\pm$ 0.22 &     0.80 $\pm$ 0.04 &     0.39 $\pm$ 0.08 &     0.36 $\pm$ 0.02 &     14.78 $\pm$ 0.96 & \bf 0.34 $\pm$ 0.00 \\ 
                                      & \ourmethod{} & \bf 10.23 $\pm$ 0.66 & \bf 0.21 $\pm$ 0.03 &     0.69 $\pm$ 0.06 & \bf 0.55 $\pm$ 0.17 & \bf 0.19 $\pm$ 0.01 & \bf 30.46 $\pm$ 0.84 &     0.33 $\pm$ 0.01 \\ 
            \hline
            \multirow{3}{*}{Apple}    & DDIM      &     19.68 $\pm$ 2.39 &     0.36 $\pm$ 0.13 &     0.63 $\pm$ 0.04 &     0.70 $\pm$ 0.12 & \bf 0.11 $\pm$ 0.00 &     36.11 $\pm$ 0.34 & \bf 0.30 $\pm$ 0.00 \\ 
                                      & CADS      &     27.83 $\pm$ 1.34 &     0.83 $\pm$ 0.11 &     0.48 $\pm$ 0.02 &     0.71 $\pm$ 0.12 & \bf 0.11 $\pm$ 0.01 &     36.65 $\pm$ 0.75 &     0.29 $\pm$ 0.00 \\ 
                                      & \ourmethod{} & \bf 14.55 $\pm$ 1.49 & \bf 0.12 $\pm$ 0.03 & \bf 0.64 $\pm$ 0.05 & \bf 0.78 $\pm$ 0.10 & \bf 0.11 $\pm$ 0.01 & \bf 36.70 $\pm$ 0.63 & \bf 0.30 $\pm$ 0.00 \\ 
            \hline
            \multirow{3}{*}{Burberry} & DDIM      &     10.43 $\pm$ 0.58 &     0.37 $\pm$ 0.02 & \bf 0.75 $\pm$ 0.04 &     0.73 $\pm$ 0.05 &     0.30 $\pm$ 0.01 &     16.80 $\pm$ 0.45 & \bf 0.35 $\pm$ 0.00 \\ 
                                      & CADS      &     12.31 $\pm$ 0.69 &     0.46 $\pm$ 0.04 &     0.69 $\pm$ 0.03 &     0.69 $\pm$ 0.04 &     0.30 $\pm$ 0.01 &     16.70 $\pm$ 0.68 & \bf 0.35 $\pm$ 0.00 \\ 
                                      & \ourmethod{} & \bf  9.61 $\pm$ 0.51 & \bf 0.27 $\pm$ 0.04 &     0.69 $\pm$ 0.04 & \bf 0.89 $\pm$ 0.07 & \bf 0.21 $\pm$ 0.01 & \bf 23.31 $\pm$ 0.56 & \bf 0.35 $\pm$ 0.00 \\ 
            \hline
            \multirow{3}{*}{Frank}    & DDIM      &      8.76 $\pm$ 0.39 &     1.31 $\pm$ 0.11 &     0.99 $\pm$ 0.01 &     0.52 $\pm$ 0.05 &     0.23 $\pm$ 0.00 &     20.81 $\pm$ 0.55 & \bf 0.29 $\pm$ 0.00 \\ 
                                      & CADS      &     10.06 $\pm$ 0.28 &     1.61 $\pm$ 0.04 & \bf 1.00 $\pm$ 0.01 &     0.55 $\pm$ 0.07 &     0.24 $\pm$ 0.01 &     21.18 $\pm$ 0.52 & \bf 0.29 $\pm$ 0.00 \\ 
                                      & \ourmethod{} & \bf  5.09 $\pm$ 0.33 & \bf 0.57 $\pm$ 0.14 &     0.93 $\pm$ 0.03 & \bf 0.62 $\pm$ 0.10 & \bf 0.16 $\pm$ 0.01 & \bf 29.41 $\pm$ 1.66 &     0.28 $\pm$ 0.00 \\ 
            \hline
            \multirow{3}{*}{Nouns}    & DDIM      &      2.88 $\pm$ 0.43 &     0.03 $\pm$ 0.00 &     0.12 $\pm$ 0.05 &     0.71 $\pm$ 0.17 &     0.48 $\pm$ 0.01 &     11.90 $\pm$ 0.44 & \bf 0.24 $\pm$ 0.00 \\ 
                                      & CADS      &      3.11 $\pm$ 0.46 &     0.03 $\pm$ 0.00 &     0.11 $\pm$ 0.02 &     0.63 $\pm$ 0.12 &     0.48 $\pm$ 0.01 &     11.78 $\pm$ 0.41 & \bf 0.24 $\pm$ 0.00 \\ 
                                      & \ourmethod{} & \bf  2.64 $\pm$ 0.31 & \bf 0.02 $\pm$ 0.00 & \bf 0.15 $\pm$ 0.05 & \bf 0.79 $\pm$ 0.11 & \bf 0.44 $\pm$ 0.01 & \bf 13.65 $\pm$ 0.45 & \bf 0.24 $\pm$ 0.00 \\ 
            \hline
            \multirow{3}{*}{Onepiece} & DDIM      &      9.17 $\pm$ 0.21 &     1.24 $\pm$ 0.03 &     0.72 $\pm$ 0.06 &     0.20 $\pm$ 0.03 &     0.29 $\pm$ 0.01 &     22.56 $\pm$ 0.41 & \bf 0.30 $\pm$ 0.00 \\ 
            ~                         & CADS      &     10.56 $\pm$ 0.99 &     1.52 $\pm$ 0.17 &     0.70 $\pm$ 0.05 &     0.19 $\pm$ 0.04 &     0.30 $\pm$ 0.01 &     22.46 $\pm$ 0.51 & \bf 0.30 $\pm$ 0.00 \\ 
            ~                         & \ourmethod{} & \bf  5.57 $\pm$ 0.64 & \bf 0.58 $\pm$ 0.10 & \bf 0.73 $\pm$ 0.06 & \bf 0.42 $\pm$ 0.06 & \bf 0.25 $\pm$ 0.01 & \bf 26.67 $\pm$ 0.62 & \bf 0.30 $\pm$ 0.00 \\ 
            \hline
            \multirow{3}{*}{Pokemon}  & DDIM      & \bf 13.23 $\pm$ 0.43 &     0.24 $\pm$ 0.02 &     0.46 $\pm$ 0.08 &     0.84 $\pm$ 0.07 &     0.28 $\pm$ 0.01 &     21.94 $\pm$ 0.58 & \bf 0.31 $\pm$ 0.00 \\ 
            ~                         & CADS      &     17.05 $\pm$ 0.40 &     0.29 $\pm$ 0.04 &     0.44 $\pm$ 0.03 &     0.76 $\pm$ 0.12 &     0.29 $\pm$ 0.01 &     21.98 $\pm$ 0.76 & \bf 0.31 $\pm$ 0.00 \\ 
            ~                         & \ourmethod{} &     13.54 $\pm$ 0.63 & \bf 0.22 $\pm$ 0.01 & \bf 0.47 $\pm$ 0.09 & \bf 0.88 $\pm$ 0.07 & \bf 0.23 $\pm$ 0.00 & \bf 23.13 $\pm$ 0.22 & \bf 0.31 $\pm$ 0.00 \\ 
            \hline
            \multirow{3}{*}{Rococo}   & DDIM      &     29.56 $\pm$ 1.66 &     7.01 $\pm$ 0.51 &     0.93 $\pm$ 0.03 &     0.45 $\pm$ 0.07 &     0.20 $\pm$ 0.00 &     19.09 $\pm$ 0.32 &     0.30 $\pm$ 0.00 \\ 
                                      & CADS      &     33.52 $\pm$ 0.63 &     8.13 $\pm$ 0.18 & \bf 0.94 $\pm$ 0.02 &     0.52 $\pm$ 0.09 &     0.21 $\pm$ 0.00 &     18.74 $\pm$ 0.57 &     0.30 $\pm$ 0.00 \\ 
                                      & \ourmethod{} & \bf 23.37 $\pm$ 0.36 & \bf 5.07 $\pm$ 0.24 & \bf 0.94 $\pm$ 0.03 & \bf 0.80 $\pm$ 0.04 & \bf 0.17 $\pm$ 0.01 & \bf 33.74 $\pm$ 0.82 & \bf 0.31 $\pm$ 0.00 \\ 
            \hline
        \end{tabular}
    }
    \label{tab:normal_train5}
\end{table}

\begin{table}
    \centering
    \caption{\looseness=-10000 Quantitative comparison between DDIM, CADS, and \ourmethod{} applied on standard fine-tuning checkpoints for $25$-shot learning on various generative modeling metrics. We show the mean and standard deviation values over 5 repeated runs in each cell.}
    \resizebox{\textwidth}{!}{
        \begin{tabular}{|c|c|c|c|c|c|c|c|c|}
        \hline
            \bf Subset &  \bf Method &  \bf FID $\downarrow$ &  \bf KID $\downarrow$ &  \bf Precision $\uparrow$ &  \bf Recall $\uparrow$ &  \bf MSS $\downarrow$ &  \bf Vendi $\uparrow$ &  \bf Prompt Fid. $\uparrow$ \\ 
            \hline
            \multirow{3}{*}{Amedeo}   & DDIM      &     14.87 $\pm$ 1.42 &     3.01 $\pm$ 0.37 & \bf 0.66 $\pm$ 0.07 &     0.69 $\pm$ 0.19 &     0.30 $\pm$ 0.01 &     14.07 $\pm$ 0.41 & \bf 0.34 $\pm$ 0.00 \\
                                      & CADS      &     16.51 $\pm$ 0.98 &     3.48 $\pm$ 0.32 &     0.58 $\pm$ 0.08 &     0.57 $\pm$ 0.19 &     0.32 $\pm$ 0.01 &     13.68 $\pm$ 0.26 & \bf 0.34 $\pm$ 0.00 \\ 
                                      & \ourmethod{} & \bf 12.22 $\pm$ 1.04 & \bf 2.89 $\pm$ 0.24 &     0.41 $\pm$ 0.07 & \bf 0.75 $\pm$ 0.05 & \bf 0.29 $\pm$ 0.01 & \bf 16.03 $\pm$ 0.30 &     0.33 $\pm$ 0.01 \\ 
            \hline
            \multirow{3}{*}{Apple}    & DDIM      &     45.35 $\pm$ 2.07 &     2.77 $\pm$ 0.22 & \bf 0.58 $\pm$ 0.03 &     0.76 $\pm$ 0.08 &     0.18 $\pm$ 0.01 &     17.85 $\pm$ 0.54 & \bf 0.31 $\pm$ 0.00 \\ 
                                      & CADS      &     30.89 $\pm$ 0.35 &     2.73 $\pm$ 0.09 & \bf 0.58 $\pm$ 0.08 &     0.78 $\pm$ 0.05 &     0.19 $\pm$ 0.01 &     17.13 $\pm$ 0.52 & \bf 0.31 $\pm$ 0.00 \\ 
                                      & \ourmethod{} & \bf 25.16 $\pm$ 1.07 & \bf 2.48 $\pm$ 0.14 & \bf 0.58 $\pm$ 0.05 & \bf 0.80 $\pm$ 0.08 & \bf 0.14 $\pm$ 0.01 & \bf 20.01 $\pm$ 0.64 & \bf 0.31 $\pm$ 0.00 \\ 
            \hline
            \multirow{3}{*}{Burberry} & DDIM      &     16.44 $\pm$ 0.18 &     1.39 $\pm$ 0.19 & \bf 0.95 $\pm$ 0.02 &     0.82 $\pm$ 0.12 &     0.27 $\pm$ 0.01 &     13.76 $\pm$ 0.42 & \bf 0.34 $\pm$ 0.00 \\ 
                                      & CADS      &     17.87 $\pm$ 1.91 &     1.51 $\pm$ 0.25 &     0.94 $\pm$ 0.03 &     0.73 $\pm$ 0.09 &     0.26 $\pm$ 0.01 &     14.06 $\pm$ 0.42 & \bf 0.34 $\pm$ 0.00 \\ 
                                      & \ourmethod{} & \bf  6.72 $\pm$ 1.84 & \bf 0.82 $\pm$ 0.26 &     0.89 $\pm$ 0.07 & \bf 0.93 $\pm$ 0.07 & \bf 0.16 $\pm$ 0.01 & \bf 20.38 $\pm$ 0.61 & \bf 0.34 $\pm$ 0.00 \\ 
            \hline
            \multirow{3}{*}{Frank}    & DDIM      &      6.13 $\pm$ 0.28 &     0.62 $\pm$ 0.09 & \bf 0.98 $\pm$ 0.00 &     0.81 $\pm$ 0.04 & \bf 0.14 $\pm$ 0.00 &     19.77 $\pm$ 0.47 & \bf 0.31 $\pm$ 0.00 \\ 
                                      & CADS      & \bf  4.57 $\pm$ 0.10 &     0.60 $\pm$ 0.07 &     0.96 $\pm$ 0.00 &     0.82 $\pm$ 0.04 & \bf 0.14 $\pm$ 0.00 &     19.50 $\pm$ 0.21 & \bf 0.31 $\pm$ 0.00 \\ 
                                      & \ourmethod{} &      4.90 $\pm$ 0.19 & \bf 0.34 $\pm$ 0.02 &     0.93 $\pm$ 0.09 & \bf 0.83 $\pm$ 0.03 & \bf 0.14 $\pm$ 0.00 & \bf 21.73 $\pm$ 0.26 & \bf 0.31 $\pm$ 0.01 \\ 
            \hline
            \multirow{3}{*}{Nouns}    & DDIM      &      1.81 $\pm$ 0.09 &     0.02 $\pm$ 0.00 &     0.59 $\pm$ 0.09 &     0.95 $\pm$ 0.05 &     0.50 $\pm$ 0.01 &      8.58 $\pm$ 0.14 & \bf 0.25 $\pm$ 0.00 \\ 
                                      & CADS      &      1.24 $\pm$ 0.05 &     0.02 $\pm$ 0.00 & \bf 0.66 $\pm$ 0.12 &     0.97 $\pm$ 0.05 &     0.51 $\pm$ 0.01 &      8.34 $\pm$ 0.18 & \bf 0.25 $\pm$ 0.00 \\ 
                                      & \ourmethod{} & \bf  1.22 $\pm$ 0.01 & \bf 0.01 $\pm$ 0.00 &     0.62 $\pm$ 0.08 & \bf 0.98 $\pm$ 0.03 & \bf 0.46 $\pm$ 0.01 & \bf 14.55 $\pm$ 0.19 & \bf 0.25 $\pm$ 0.00 \\ 
            \hline
            \multirow{3}{*}{Onepiece} & DDIM      &      2.85 $\pm$ 0.13 &     0.10 $\pm$ 0.02 & \bf 0.83 $\pm$ 0.04 &     0.69 $\pm$ 0.05 & \bf 0.25 $\pm$ 0.00 &     16.86 $\pm$ 0.15 &     0.30 $\pm$ 0.00 \\ 
            ~                         & CADS      &      3.02 $\pm$ 0.16 &     0.15 $\pm$ 0.01 &     0.78 $\pm$ 0.03 &     0.75 $\pm$ 0.02 &     0.26 $\pm$ 0.01 &     16.81 $\pm$ 0.29 & \bf 0.31 $\pm$ 0.00 \\ 
            ~                         & \ourmethod{} & \bf  2.14 $\pm$ 0.11 & \bf 0.09 $\pm$ 0.01 &     0.82 $\pm$ 0.04 & \bf 0.79 $\pm$ 0.01 & \bf 0.25 $\pm$ 0.01 & \bf 18.05 $\pm$ 0.45 &     0.30 $\pm$ 0.00 \\ 
            \hline
            \multirow{3}{*}{Pokemon}  & DDIM      &      6.32 $\pm$ 0.34 & \bf 0.05 $\pm$ 0.01 &     0.78 $\pm$ 0.05 &     0.78 $\pm$ 0.05 &     0.31 $\pm$ 0.01 &     14.89 $\pm$ 0.44 & \bf 0.31 $\pm$ 0.00 \\ 
            ~                         & CADS      &      8.20 $\pm$ 0.30 &     0.06 $\pm$ 0.00 & \bf 0.86 $\pm$ 0.03 &     0.74 $\pm$ 0.03 &     0.31 $\pm$ 0.02 &     14.77 $\pm$ 0.67 & \bf 0.31 $\pm$ 0.00 \\ 
            ~                         & \ourmethod{} & \bf  4.88 $\pm$ 0.93 & \bf 0.05 $\pm$ 0.00 &     0.82 $\pm$ 0.05 & \bf 0.86 $\pm$ 0.03 & \bf 0.28 $\pm$ 0.01 & \bf 15.95 $\pm$ 0.42 & \bf 0.31 $\pm$ 0.00 \\ 
            \hline
            \multirow{3}{*}{Rococo}   & DDIM      &     10.30 $\pm$ 0.58 &     1.82 $\pm$ 0.09 &     0.83 $\pm$ 0.02 &     0.94 $\pm$ 0.02 &     0.15 $\pm$ 0.00 &     15.08 $\pm$ 0.37 & \bf 0.34 $\pm$ 0.00 \\ 
                                      & CADS      &     13.36 $\pm$ 0.61 &     1.85 $\pm$ 0.11 &     0.83 $\pm$ 0.02 &     0.92 $\pm$ 0.00 &     0.15 $\pm$ 0.00 &     15.15 $\pm$ 0.34 & \bf 0.34 $\pm$ 0.00 \\ 
                                      & \ourmethod{} & \bf  9.41 $\pm$ 1.06 & \bf 1.63 $\pm$ 0.07 & \bf 0.86 $\pm$ 0.03 & \bf 0.97 $\pm$ 0.01 & \bf 0.14 $\pm$ 0.01 & \bf 15.89 $\pm$ 0.84 & \bf 0.34 $\pm$ 0.00 \\ 
            \hline
        \end{tabular}
    }
    \label{tab:normal_train25}
\end{table}

\paragraph{Varying the Number of Training Samples.}
\looseness=-10000
We conduct a quantitative evaluation for model checkpoints that are trained on 5 and 25 samples of each FSCG-8 subset respectively. In both cases, the models are trained for 2000 iterations. For the 5-shot fine-tuning runs, there are 45 validation samples, while for the 25-shot fine-tuning runs, there are 25 validation samples. We generate the same number of images as the number of validation samples in each evaluation run. The results are presented in Table \ref{tab:normal_train5} and \ref{tab:normal_train25}. As shown in the tables, \ourmethod{} achieves the best performance in terms of matching the validation distribution (FID, KID) and generating diverse samples (Recall, MSS, Vendi Score). \ourmethod{} also remains competitive in quality-focused metrics (Precision and Prompt Fidelity).

\begin{figure}[!h]
  \centering
  \includegraphics[trim={0cm 0cm 0cm 0cm},width=\textwidth,center]
  {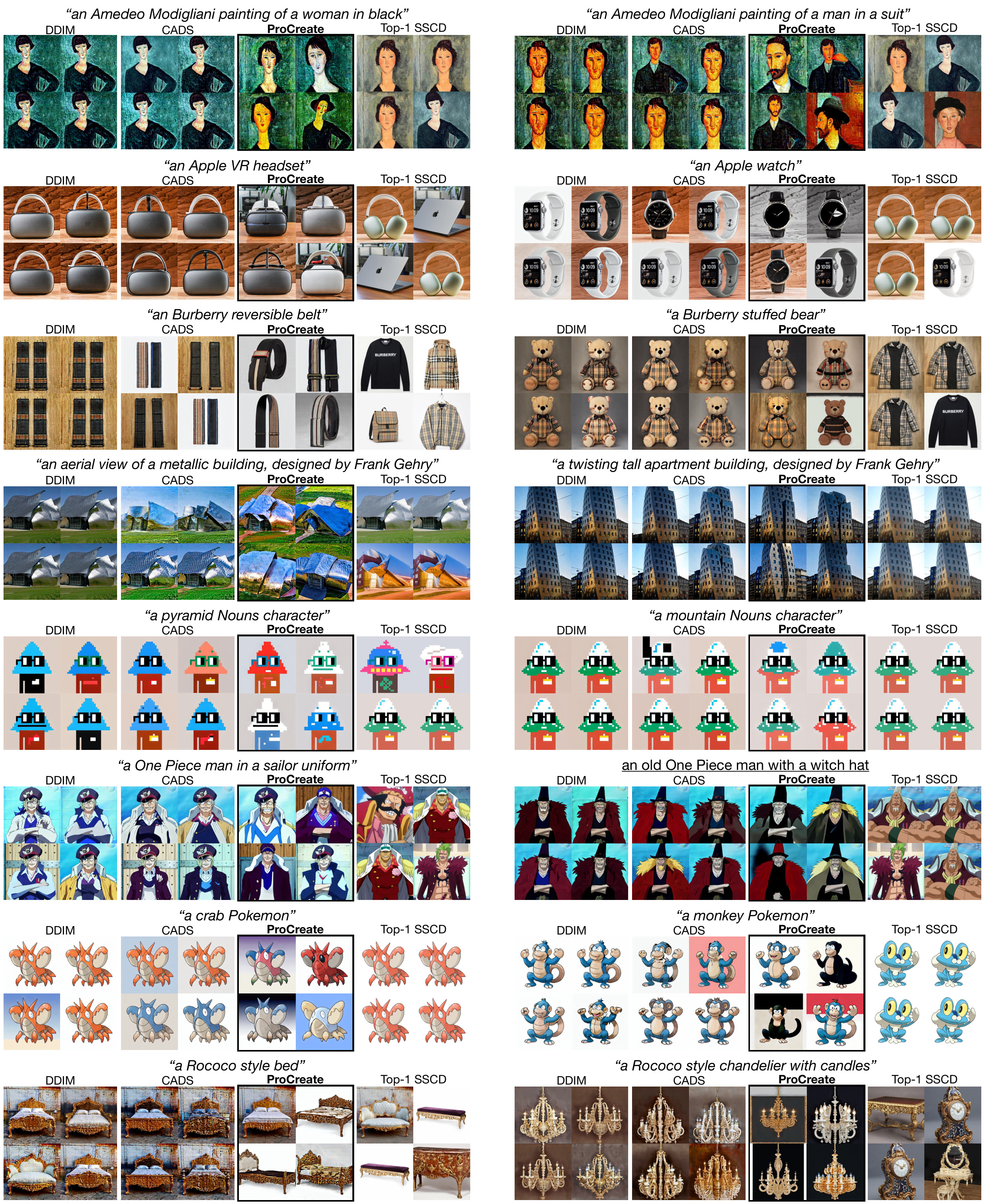}
  \caption{\looseness=-10000 Qualitative comparison between DDIM, CADS, and \ourmethod{} for few-shot creative generation on \ourdataset{} with DreamBooth fine-tuning. For each sampling method, we show two prompts and four generated samples for each prompt. In addition, we match each sample from \ourmethod{} with its closest training image based on the SSCD score \cite{sscd} between the matched pair.}
  \label{fig:dreambooth_train10_qualitative}
\end{figure}

\paragraph{DreamBooth Qualitative Samples.}
\looseness=-10000
In Figure \ref{fig:dreambooth_train10_qualitative}, we perform DreamBooth fine-tuning with the experiment setup from Section \ref{sec:fscg_experiment_setup} to produce qualitative results with DDIM, CADS, and \ourmethod{} sampling methods. Similarly to the qualitative comparisons in Figure \ref{fig:normal_train10_qualitative}, \ourmethod{} generates the most diverse samples that follow the prompts yet do not replicate the Top-1 SSCD matched training images.

\end{document}